\documentclass[10pt,journal,compsoc]{IEEEtran}
\ifCLASSINFOpdf
\else
\fi

\usepackage{graphicx}
\usepackage{amsmath,amssymb}
\usepackage{booktabs}
\usepackage{array}
\usepackage{multirow}
\usepackage{makecell}
\usepackage{float}
\usepackage{bm}
\newcommand{\best}[1]{\textbf{\boldmath $#1$}}
\usepackage{tikz}
\usetikzlibrary{positioning,arrows.meta}

\tikzset{
    box/.style={draw, rounded corners, align=center, minimum width=2.2cm, minimum height=0.9cm, font=\footnotesize},
    smallbox/.style={draw, rounded corners, align=center, minimum width=1.9cm, minimum height=0.8cm, font=\footnotesize},
    arrow/.style={->, thick}
}

\begin{document}
%
\title{A Systematic Failure Analysis of Vision Foundation Models for Open Set Iris Presentation Attack Detection}


\author{\IEEEauthorblockN{Rahul Anand\IEEEauthorrefmark{1} \IEEEauthorrefmark{3},
Siddharth Singh\IEEEauthorrefmark{1} \IEEEauthorrefmark{3},
Dileep A D\IEEEauthorrefmark{1}, 
Mahadeva Prasanna\IEEEauthorrefmark{2}, and
Raghavendra Ramachandra\IEEEauthorrefmark{3}}
\IEEEauthorblockA{\IEEEauthorrefmark{1}Indian Institute of Technology, Dharwad, India. \\}
\IEEEauthorblockA{\IEEEauthorrefmark{2}Indian Institute of Information Technology Dharwad, India.\\}
\IEEEauthorblockA{\IEEEauthorrefmark{3}SAFE Center, Norwegian University of Science and Technology (NTNU), Norway.}

\thanks{Manuscript received December XX, XXXX; revised August XX, XX. 
Corresponding author: Raghavendra Ramachandra (email: raghavendra.ramachandra@ntnu.no).\\
Rahul Anand, Siddharth Singh and Raghavendra Ramachandra contributed equally to this work.}}

\markboth{Preprint: Accepted in IEEE Transactions on Biometrics, Behavior, and Identity Science (T-BIOM) }%
{Shell \MakeLowercase{\textit{et al.}}: Bare Demo of IEEEtran.cls for IEEE Transactions on Magnetics Journals}
%



\IEEEtitleabstractindextext{%
\begin{abstract}
Vision foundation models have demonstrated strong transferability across diverse visual recognition tasks and are increasingly considered for biometric applications. Their suitability for iris Presentation Attack Detection (PAD), particularly under realistic open-set operating conditions, remains insufficiently examined. This work presents a systematic failure analysis of general-purpose vision foundation models for open-set iris PAD using periocular imagery. Five representative foundation models are evaluated under three open-set protocols that explicitly separate different sources of distribution shift: (i) unseen Presentation Attack Instruments (PAIs), (ii) unseen datasets captured with different sensors, and (iii) cross-spectral transfer from near-infrared (NIR) to visible spectrum (VIS) imagery. Both frozen feature representations and parameter-efficient task adaptation using Low-Rank Adaptation (LoRA) are assessed within a unified experimental framework.
The results indicate that foundation models can transfer across datasets with similar sensing characteristics, but fail to generalise reliably to unseen attack instruments and degrade sharply under cross-spectral evaluation. While LoRA improves performance in certain cross-dataset settings, it frequently amplifies failure under attack-level and spectral shifts. Additional validation experiments using segmented iris inputs, full backbone fine-tuning, joint cross-dataset and cross-PAI shifts, and reverse VIS$\rightarrow$NIR transfer further confirm that these failures are not simply artefacts of periocular input, weak adaptation, or one-directional spectral evaluation. These findings show that strong closed-set or cross-dataset performance should not be treated as evidence of robust open-set security, and highlight the need for PAD representations that maintain sensitivity to presentation artefacts while remaining stable under realistic deployment variation.
\end{abstract}

}

\maketitle

\IEEEdisplaynontitleabstractindextext

%
\IEEEpeerreviewmaketitle

\section{Introduction}
Iris recognition is widely deployed in high-security applications owing to its high distinctiveness and long-term stability. However, as with other biometric modalities, iris recognition systems are vulnerable to presentation attacks, in which an adversary presents an artificial or altered iris sample to the sensor with the intention of gaining unauthorised access. Common Presentation Attack Instruments (PAIs) include printed iris images, textured contact lenses, electronic display attacks, and, more recently, synthetic irises generated using learning-based synthesis. The detection of such attacks, commonly referred to as iris PAD, has therefore become an essential component of secure iris biometric systems \cite{morales2023introduction,yambay2023review}.

Substantial progress has been achieved in iris PAD over the past decade. Both handcrafted feature-based approaches and deep learning methods have demonstrated strong performance under controlled evaluation settings, particularly in closed-set scenarios where the attack types, sensors, and datasets observed during testing are also represented during training \cite{nguyen2024deep}. Large-scale evaluation campaigns and competitions have further contributed to this progress by standardising protocols and enabling comparison of methods under fixed conditions \cite{yambay2023review}. As a consequence, many contemporary systems achieve near-perfect accuracy when evaluated in closed-set settings.

Closed-set evaluation, however, does not accurately reflect the conditions encountered in operational deployments. In practice, iris PAD systems are frequently exposed to PAIs not represented during training, to data acquired from previously unseen sensors, and to images captured under different spectral conditions. The ability to generalise to such previously unseen conditions is therefore critical for practical security. Recent studies have shown that performance may degrade substantially when PAD models are evaluated under open-set conditions in which one or more factors differ between training and testing \cite{boyd2023comprehensive}. These observations highlight that strong closed-set performance does not guarantee robust protection under real-world deployment scenarios.

Open-set iris PAD has consequently emerged as an important research problem. Prior work has examined cross-sensor, cross-dataset, and cross-attack evaluations, consistently demonstrating that generalisation remains challenging even for modern deep learning approaches \cite{boyd2023comprehensive}. In parallel, vision foundation models trained on large-scale data using self-supervised or weakly supervised objectives have shown strong transferability across diverse visual recognition tasks. Their growing adoption in biometric applications naturally raises the question of whether such models can mitigate the generalisation limitations observed in iris PAD, particularly under open-set conditions.


\vspace{0.2cm}
\subsection{Contributions and Research Questions}

This work presents a systematic analysis of the behaviour of vision foundation models for iris PAD under open-set conditions. Rather than proposing a new detection architecture, the focus is on understanding the generalisation limits of existing large-scale visual representations when exposed to previously unseen attacks, datasets, and sensing conditions. The main contributions of this study are summarised as follows:

\begin{itemize}
\item We conduct a large-scale evaluation of five vision foundation models for iris PAD using a unified experimental framework and a diverse corpus of near-infrared (NIR) and visible spectrum (VIS) iris imagery.
\item We design and apply three open-set evaluation protocols that isolate distinct sources of distribution shift, namely unseen PAIs, unseen datasets captured with different sensors, and cross-spectral transfer from NIR to VIS imagery.
\item We analyse the impact of parameter-efficient task adaptation via Low-Rank Adaptation (LoRA) on open-set generalisation, enabling controlled comparison between frozen and adapted representations.
\item We provide an empirical failure analysis identifying the conditions under which foundation models succeed or break down in iris PAD, supported by quantitative metrics.
\item To the best of our knowledge, this is the first work that systematically dissects foundation model behaviour in iris PAD using three complementary open-set protocols, large-scale multi-source data, careful statistical evaluation, and novel geometric separability analysis of feature manifolds.
\end{itemize}

To structure the analysis and guide interpretation of results, the study addresses the following research questions:

\begin{itemize}
\item \textbf{RQ1:} Do vision foundation models generalise to previously unseen presentation attack instruments in iris PAD?
\item \textbf{RQ2:} Are foundation model representations robust to dataset-level shifts when the sensing modality remains unchanged?
\item \textbf{RQ3:} Do vision foundation models learn features that transfer across imaging spectra for iris PAD?
\item \textbf{RQ4:} Does parameter-efficient fine-tuning mitigate or amplify open-set failure in iris PAD?
\end{itemize}

In this work, we investigate these questions through a systematic evaluation of vision foundation models for open-set iris PAD using periocular imagery. While recent studies have begun exploring foundation models in closed-set or limited biometric PAD contexts, their behaviour under realistic open-set conditions, particularly the degree of failure across isolated distribution shifts and the counter-intuitive impact of modern adaptation techniques, remains largely unexamined. Rather than focusing solely on closed-set performance, we analyse their behaviour under cross-PAI, cross-dataset, and cross-spectral protocols that reflect increasingly realistic deployment scenarios. To the best of our knowledge, this is the first large-scale, protocol-driven failure analysis of current vision foundation models in iris PAD, revealing previously undocumented vulnerabilities and guiding the development of more robust, PAD-aware representations. Our objective is to assess the extent to which foundation models improve open-set iris PAD and to identify the conditions under which their representations remain insufficient for reliable iris PAD.

The remainder of the paper is organised as follows. Section~2 reviews related work. Section~3 describes the foundation-model evaluation framework. Section~4 introduces the datasets and open-set protocols. Section~5 presents the main PAD results, Section~6 analyses feature-space separability, Section~7 discusses implications for PAD design, Section~8 reports targeted extended validations, and Section~9 concludes the paper.
\section{Related Work}

This section situates the present study with respect to prior work on iris PAD. We first review handcrafted and deep learning-based approaches, followed by studies addressing open-set and cross-domain evaluation. Finally, we discuss the recent use of vision foundation models in biometrics and outline the gaps addressed in this work.

\subsection{Classical and Deep Learning-based Iris PAD}

Early research in iris PAD concentrated on handcrafted feature representations designed to capture texture inconsistencies introduced by presentation attacks. Methods based on multiscale texture descriptors, local binary patterns, and binarised statistical image features were widely examined. Raghavendra and Busch proposed a multiscale binarised statistical image feature model demonstrating strong performance against printed and textured contact lens attacks under controlled settings~\cite{raghavendra2015robust}. Subsequent work extended handcrafted approaches via feature fusion strategies that combined complementary texture and frequency cues~\cite{choudhary2020iris}. Other studies targeted local artefacts such as micro-stripe patterns or saliency-guided regions to detect textured contact lenses~\cite{fang2020micro,parzianello2022saliency}. A comprehensive overview of handcrafted iris PAD techniques is provided by Morales \emph{et al.}~\cite{morales2023introduction}.

With the availability of annotated datasets, deep learning-based approaches became dominant. Convolutional neural networks were trained end-to-end or used as feature extractors to distinguish bona fide samples from attacks. Architectures such as AlexNet variants, DenseNet-based models, and task-specific convolutional designs demonstrated strong performance in controlled settings~\cite{yadav2019detecting,balashanmugam2023effective}. More recent work introduced supervised representation learning strategies that explicitly promote class separability for PAD, improving discrimination for known attack types~\cite{gautam2022deep}. Survey studies confirm that deep learning-based methods now outperform handcrafted techniques in closed-set evaluations, but also note their sensitivity to dataset bias and attack variability~\cite{nguyen2024deep}.

Despite their success, both handcrafted and deep learning-based PAD methods typically assume that attack characteristics observed during testing are adequately represented during training. Under such assumptions, closed-set performance may be high; however, models often degrade sharply when evaluated under open-set conditions involving previously unseen PAIs, sensors, or acquisition characteristics.

\subsection{Open-Set and Cross-Domain Iris PAD}

To bridge the gap between laboratory evaluation and deployment, several studies have examined iris PAD under open-set and cross-domain conditions \cite{boyd2023comprehensive, hoffman2018convolutional, Pal}. These include cross-sensor, cross-dataset, and cross-attack evaluations in which one or more factors differ between training and testing. Such protocols aim to assess the ability of PAD systems to generalise beyond closed-set assumptions.

Boyd \emph{et al.}\ conducted a comprehensive study on open-set iris PAD, systematically analysing generalisation to unseen PAIs and unseen datasets~\cite{boyd2023comprehensive}. Their findings showed that even state-of-the-art systems experienced substantial degradation when exposed to such shifts. Similar behaviour has been observed in LivDet evaluation campaigns, where algorithms that perform well on known attacks frequently fail to generalise to novel PAIs or new sensors~\cite{yambay2023review}. These studies indicate that open-set iris PAD remains unsolved, with generalisation limited across attack types and acquisition conditions.

Most existing work on open-set PAD evaluates specific architectures or handcrafted pipelines. In contrast, little attention has been given to how large-scale pretrained visual representations behave under these shifts or whether such representations mitigate open-set failure.

\subsection{Foundation Models in Biometrics}

Vision foundation models trained on large-scale data using self-supervised or weakly supervised objectives have recently gained attention within the biometrics community. Their cross-domain transferability has motivated applications in face recognition, fingerprint analysis, and PAD. Surveys on foundation models for biometrics provide an overview of their potential benefits and limitations, particularly with respect to data efficiency and representation reuse~\cite{shahreza2025foundation}.

Benchmarking efforts have evaluated foundation models for biometric recognition tasks, including zero-shot and few-shot scenarios, and have reported promising performance for identity-related problems~\cite{sony2025benchmarking}. In the specific context of iris PAD, Tapia \emph{et al.}\ fine-tuned foundation models and demonstrated competitive performance under closed-set evaluation~\cite{tapia2025towards}. More recently, SpectraIrisPAD explored foundation-model adaptation for multispectral iris PAD and reported improvements in controlled multispectral settings~\cite{ramachandra2025spectrairispad}. These studies indicate that foundation models can be useful for iris PAD when the evaluation setting is closed-set, controlled, or explicitly adapted to the sensing condition.

The present work addresses a different but complementary question. Rather than proposing a specialised PAD model, we analyse whether general-purpose vision foundation models retain reliable PAD behaviour under open-set shifts. In particular, we separate attack-level, dataset-level, and spectral shifts to examine whether performance under one source of variation transfers to another. Our results show that favourable transfer under dataset-level variation does not necessarily imply robustness to unseen PAIs or spectral mismatch. Thus, this work complements recent foundation-model-based iris PAD studies by identifying deployment conditions under which general visual representations fail, and by providing protocol-driven evidence for the need for PAD-aware adaptation.

Existing studies do not yet provide a systematic failure analysis of general-purpose foundation-model representations for iris PAD under practical open-set conditions. In particular, generalisation to unseen PAIs, transfer across imaging spectra, and the effect of adaptation under isolated distribution shifts remain insufficiently understood. The present study addresses these gaps through a controlled, protocol-driven analysis of failure modes, thereby uncovering key vulnerabilities and offering practical guidance for future PAD-aware adaptations.

\section{Foundation Models for Iris PAD}
This section introduces the vision foundation models analysed in this study and explains how they are employed for iris PAD using periocular imagery. The focus is not on architectural novelty, but on characterising the behaviour of existing large-scale visual representations under open-set conditions.

\begin{figure*}[htp]
\centering
\includegraphics[width=0.95\textwidth]{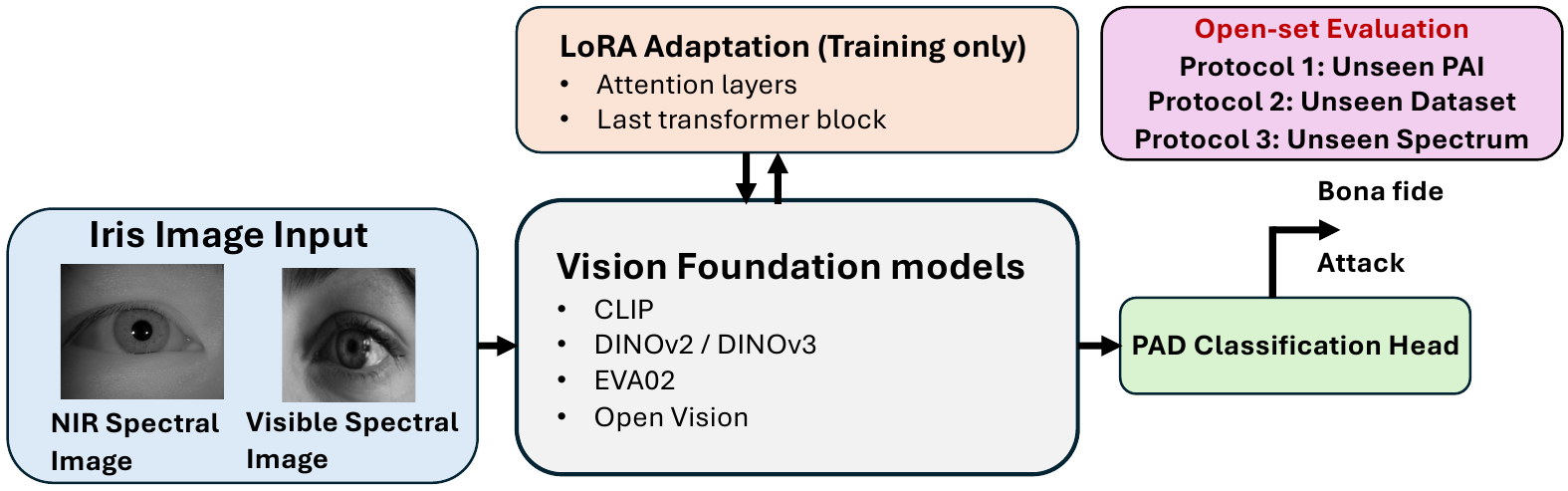}
\caption{Conceptual overview of open-set iris PAD using periocular images and vision foundation models. Models are trained under known conditions and evaluated under open-set scenarios involving unseen presentation attack instruments, datasets, and imaging spectra.}
\label{fig:proposed}
\end{figure*}

\subsection{Vision Foundation Models}

Vision foundation models are large-scale neural networks pretrained on extensive image corpora using self-supervised or weakly supervised objectives. Such pretraining aims to produce broadly transferable visual representations for downstream tasks. In this work, we evaluate five representative foundation models that have recently gained attention in biometric research: CLIP~\cite{radford2021learning}, DINOv2~\cite{oquab2023dinov2}, DINOv3~\cite{caron2024dinov3}, EVA02~\cite{fang2024eva02}, and OpenVision~\cite{li2025openvision}. Note that the large versions of each of these models have been used during evaluation.

These models differ in their pretraining objectives. CLIP~\cite{radford2021learning} learns image representations via contrastive alignment with text, resulting in high-level semantic embeddings. DINOv2 and DINOv3~\cite{oquab2023dinov2,caron2024dinov3} employ self-distillation across augmented views, encouraging consistency and invariance in the learned features. EVA02~\cite{fang2024eva02} applies masked image modelling, reconstructing missing image regions to capture spatial structure. OpenVision~\cite{li2025openvision} provides a general-purpose transformer-based encoder for visual feature extraction.

Although these models vary in their pretraining strategies, they share a common objective: to suppress nuisance variation such as illumination shifts, sensor noise, and minor texture changes. Such invariances are advantageous for recognition tasks, where intra-class variation should be reduced. In PAD, however, many discriminative cues are subtle artefacts introduced by PAIs, often local or fine-grained in nature. Representations optimised for semantic abstraction may suppress these cues, potentially hindering generalisation under open-set shifts. This motivates a systematic evaluation of foundation models in iris PAD, rather than assuming transferability from conventional vision benchmarks.

\subsection{Use of Periocular Images}
\label{sec:periocular_input}

Traditional iris PAD pipelines often incorporate iris segmentation and normalisation before classification. In the main experiments of this study, we instead use periocular images for all NIR and VIS datasets. This choice provides a common input representation across datasets with different sensors, resolutions, spectra, and attack types, and avoids making the main protocol comparisons dependent on dataset-specific segmentation quality or normalisation procedures.

This design is not intended to suggest that iris localisation is unnecessary for PAD. Rather, it allows the behaviour of the foundation-model representations to be compared under a uniform preprocessing pipeline. Periocular-based PAD has also been considered in prior iris PAD evaluations~\cite{livdet2013,livdet2015,livdet2017,boyd2023comprehensive}. Since segmentation may remove periocular context while reducing background and eyelid variation, we explicitly evaluate this design choice in Section~\ref{sec:extended_segmentation}, where segmented iris inputs are tested under representative conditions from all three protocols.

\subsection{Fine-Tuning Strategies}

Two fine-tuning strategies are considered to evaluate the effect of task adaptation on open-set generalisation. In the first strategy, the foundation model backbone remains frozen and a lightweight MLP classifier is trained to discriminate between bona fide and attack samples. The MLP consists of two fully connected layers with a hidden dimension of 512 and a ReLU activation. This setting assesses the direct transferability of pretrained representations without modifying the embedding geometry of the backbone.

In the second strategy, the MLP head is trained jointly with parameter-efficient task adaptation using Low-Rank Adaptation (LoRA). LoRA modules are inserted into selected attention layers in the final transformer blocks, while all remaining backbone parameters are frozen. This strategy introduces a small number of trainable parameters relative to full fine-tuning, enabling a controlled comparison between frozen and adapted representations under identical training and evaluation protocols. LoRA is used solely as an analytical instrument for probing how mild adaptation affects open-set behaviour, rather than as a proposed improvement to iris PAD.

\subsection{Implementation Details}

All fine-tuning experiments are performed on a single NVIDIA L40S GPU. Training uses an 80/20 train–validation split that preserves class proportions. Models are trained for 25 epochs with batch size 64 using cross-entropy loss and optimised with Adam (learning rate $10^{-3}$, $\beta_1=0.9$, $\beta_2=0.999$). Learning rate decay is triggered when the validation loss fails to improve for two consecutive epochs. Early stopping is applied, and the checkpoint with the minimum validation loss is selected for evaluation. Importantly, no hyper-parameter tuning is performed across protocols to avoid biasing the evaluation toward particular open-set conditions.

\subsection{Relation to Open-Set Evaluation}

The models above are evaluated under three open-set protocols that isolate distinct sources of distribution shift: unseen presentation attacks, unseen datasets acquired with different sensors, and unseen spectra. By fixing the architecture and training configurations while varying only the evaluation conditions, the analysis focuses on identifying where and why foundation model representations succeed or fail in iris PAD. This structure facilitates a direct link between empirical results and the generalisation properties of the underlying representations.

\section{Datasets and Evaluation Protocols}
\label{sec:datasets}

\begin{figure*}[t!]
\begin{center}
\includegraphics[width=0.95\linewidth]{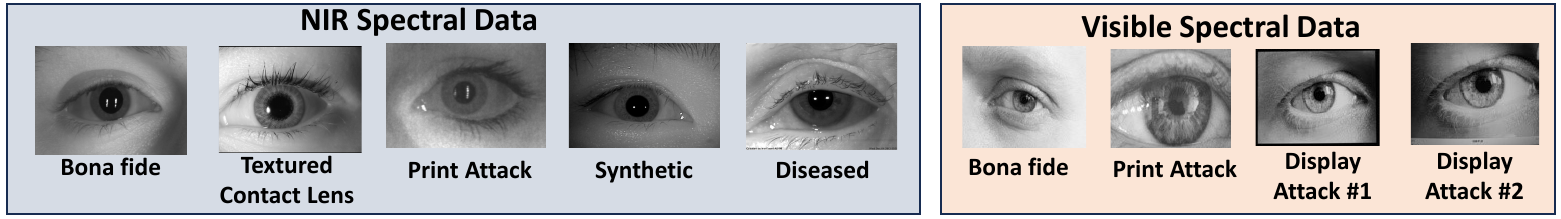}
\end{center}
   \caption{Example periocular images from the NIR (top row) and VIS (bottom row) corpora used in this study. The NIR data span four PAI categories relevant to Protocol~1: textured contact lenses, paper prints, synthetic irises, and diseased eyes. The VIS data from VSIA include bona fide samples and three intentional PAIs: print, display, and print+display. The illustration highlights both intentional spoof mechanisms and non-intentional appearance variation, motivating the need to evaluate open-set robustness beyond dataset-level shifts.}
\label{fig:db_Image}
\end{figure*}

This study uses a diverse collection of publicly available near-infrared (NIR) and visible spectrum (VIS) periocular iris datasets. The NIR corpus contains bona fide samples and four presentation attack or non-standard appearance categories: textured contact lenses, paper prints, diseased eyes, and synthetic irises. In total, the NIR corpus contains 81{,}024 images. The datasets differ in sensor type, acquisition environment, subject population, and attack characteristics, making them suitable for evaluating open-set behaviour across both dataset-level and PAI-level shifts. The distribution of samples across datasets and attack categories is provided in Table~\ref{tab:dataset-summary}. The VIS dataset used for cross-spectral evaluation is summarised in Table~\ref{tab:vsia-summary}. Figure~\ref{fig:db_Image} illustrates example iris images from the NIR and VIS corpora employed in this work.

\subsection{Near-Infrared Datasets}

\subsubsection{CASIA-IrisV4}
CASIA-IrisV4 contains ISO-compliant NIR iris images acquired under controlled imaging conditions~\cite{casia_irisv4}. In this work, bona fide samples are taken from CASIA-Iris-Thousand, while synthetically generated iris images are used to represent non-biological iris patterns~\cite{casia_iris_syn}. The original images have a resolution of $640 \times 480$ pixels and are organised into multiple subsets reflecting different acquisition settings. This dataset contributes both bona fide samples and synthetic iris attacks to the NIR corpus.

\subsubsection{IIITD-CLI}
The IIIT-Delhi Contact Lens Iris (CLI) dataset consists of NIR iris images captured using two sensors~\cite{iiitd_cli}. Each eye is recorded under three conditions: no contact lens, clear soft lens, and coloured textured lens. In this study, the no-lens samples are used as bona fide presentations, while coloured textured lens samples contribute to the textured contact lens attack category. This dataset is important for evaluating robustness to contact-lens-induced texture changes.

\subsubsection{LivDet-Iris 2013, 2015, and 2017 (Clarkson)}
The Clarkson partitions of the LivDet-Iris competitions contain bona fide NIR images together with printed iris attacks and textured contact lens attacks~\cite{livdet2013,livdet2015,livdet2017}. These datasets include multiple acquisition sensors and were collected under standardised competition protocols. They therefore provide a useful source of both sensor variation and intentional presentation attacks, and are used in the paper prints and textured contact lens categories.

\subsubsection{LivDet-Iris 2017 (Warsaw)}
The Warsaw partition of the LivDet-Iris 2017 competition includes bona fide samples and paper print attacks~\cite{livdet2017}. Images are captured using the IrisGuard AD100 sensor. This dataset contributes to the paper prints attack category and provides an additional NIR acquisition source distinct from the Clarkson partitions.

\subsubsection{NDCLD13 and NDCLD15}
The Notre Dame Contact Lens Detection datasets contain NIR iris images with and without cosmetic contact lenses~\cite{ndcld13,ndcld15}. Images are acquired using IrisGuard AD100 and LG4000 sensors and include multiple textured contact lens styles and manufacturers. These datasets are used primarily to represent textured contact lens attacks and introduce substantial variation in lens pattern, sensor type, and subject appearance.

\subsubsection{Warsaw-BioBase Disease Iris v2.1}
Warsaw-BioBase Disease Iris v2.1 contains NIR iris images from subjects exhibiting ocular pathologies such as cataract, glaucoma, or iris damage~\cite{warsaw-biobase}. Although these samples are not intentional spoofing attacks, the pathology-induced changes can substantially alter iris appearance and affect recognition reliability. We therefore treat them as a distinct non-intentional presentation variation category for open-set robustness analysis.

\subsubsection{Disease Eyes}
The Disease Eyes dataset includes NIR iris images captured before and after medical treatment of ocular conditions~\cite{disease_pre_post}. The associated pathological changes introduce appearance variation unrelated to spoofing intent, but are operationally relevant because they may cause PAD systems to encounter samples that deviate from the normal bona fide distribution. These images are incorporated into the diseased-eye category.

\subsubsection{Synthetic Iris}
The Synthetic Iris dataset contains artificially generated iris images produced using learning-based synthesis techniques~\cite{synthetic_iris}. Bona fide samples and attack samples are acquired from the Authentic noTCL and Synthetic noTCL datasets, respectively. Since the attack samples do not correspond to real subjects, they are categorised as synthetic irises. This category is used to assess whether foundation models suppress or normalise non-biological iris patterns.

\subsection{Visible Spectrum Dataset: VSIA}

The VSIA database contains VIS periocular images captured using a DSLR camera from 55 subjects~\cite{raghavendra2015robust}. It includes bona fide samples and three intentional PAI types: paper prints, electronic display attacks, and combined print-and-display attacks. Unlike the NIR corpus, VSIA is acquired in the visible spectrum and is therefore used exclusively for cross-spectral evaluation in Protocol~3. This setting enables assessment of whether models trained on NIR data can transfer to VIS imagery with different spectral characteristics and attack appearances.

\subsection{Dataset Composition and Preprocessing}
\label{sec:preprocessing}

Only single-channel NIR images are retained from the NIR datasets. Duplicate samples are removed before training. Since the original datasets differ in spatial resolution, sensor format, and eye-region framing, all NIR and VIS images are processed using a common pipeline before being passed to the foundation-model encoders. Specifically, images are converted to greyscale, cropped to a consistent periocular region, and resized to $224 \times 224$ pixels. The same preprocessing is used across Protocols~1--3 to ensure that differences in performance arise from dataset, PAI, or spectral variation rather than from protocol-specific preprocessing.

No iris segmentation or iris normalisation is applied in the main experiments. This choice avoids introducing segmentation quality as an additional confounding factor, particularly across heterogeneous NIR and VIS datasets. We acknowledge that segmentation can remove periocular context while also reducing background, eyelid, and sensor-dependent variation. Therefore, to assess whether the use of periocular imagery affects the main conclusions, Section~\ref{sec:extended_segmentation} reports an additional segmented-iris validation experiment using the same evaluation metrics and representative conditions from the main protocols.

\subsection{Semantics of Attack Categories and Evaluation Roles}

The datasets employed in this study encompass multiple sources of variation that differ in both intent and acquisition characteristics. For clarity, we distinguish between \emph{intentional} spoofs, \emph{non-intentional} appearance variations, and \emph{synthetic} iris imagery, while treating all as presentation attack instruments for the purpose of evaluating robustness under open-set conditions.

Intentional spoofs include textured contact lenses, paper prints, and electronic display attacks. These PAIs are designed to deceive iris recognition systems and constitute the canonical form of presentation attacks considered in LivDet-style evaluations. Non-intentional appearance variations arise from ocular pathologies (e.g., cataract or corneal damage). While such variations are not spoofing attempts, they introduce substantial deviations from the bona fide distribution and are known to degrade recognition performance. Their inclusion allows the evaluation of model robustness to non-standard but operationally relevant imagery. Synthetic iris images represent a third category; they are artificially generated using learning-based synthesis techniques and do not correspond to real subjects. They provide a controlled means to assess whether foundation models inadvertently normalise or suppress non-biological patterns, which is pertinent given the increasing availability of generative models.

These distinctions are not intended to redefine the taxonomy of presentation attacks but to clarify the rationale for grouping heterogeneous data sources under a unified open-set evaluation framework. In operational deployments, PAD systems must handle unexpected samples regardless of their origin or intent. Including multiple variation sources therefore enables a more comprehensive assessment of generalisation behaviour and failure modes.

\subsection{Use of Datasets in the Three Protocols}
The protocols are designed to isolate one source of distribution shift at a time. Protocol~1 isolates attack-mechanism shift, Protocol~2 isolates dataset/sensor shift, and Protocol~3 isolates spectral shift. This separation allows the observed failures to be attributed to specific open-set factors rather than to a mixture of uncontrolled variations.
\subsubsection{Protocol~1 (Cross-PAI)}
All NIR datasets are consolidated and grouped into four presentation attack instruments (PAIs): textured contact lenses, paper prints, diseased eyes, and synthetic irises. In each experimental run, one PAI and its corresponding bona fide samples are held out entirely for testing, while the remaining NIR data are used for training and validation. This protocol evaluates the ability to generalise to previously unseen attack instruments.

\subsubsection{Protocol~2 (Cross-Dataset)}
A leave-one-dataset-out evaluation is performed over the NIR datasets listed in Table~\ref{tab:dataset-summary}. For each run, a complete dataset, containing both bona fide and attack samples, is held out for testing and the remaining datasets are used for training and validation. This protocol measures generalisation across different sensors, subjects, and acquisition conditions. 

\subsubsection{Protocol~3 (Cross-Spectral)}
All NIR datasets are combined for training and validation. Testing is conducted exclusively on the VIS dataset (VSIA), which is not used for training. This protocol assesses cross-spectral transferability from NIR to VIS imagery.  

\begin{table*}[t]
\centering
\caption{NIR periocular image corpus used in this study, grouped by PAI taxonomy, dataset provenance, and sensor. ``Intentional'' denotes deliberate spoof attempts; ``Non-intentional'' denotes pathology-induced appearance variation; ``Synthetic'' denotes non-biological iris imagery.}
\label{tab:dataset-summary}
\begin{tabular}{l l l r}
\toprule
\textbf{PAI Taxonomy} & \textbf{Dataset} & \textbf{Sensor / Source} & \textbf{\# Samples} \\
\midrule

\multirow{9}{*}{\textbf{Bona fide}}
 & CASIA-IrisV4 & IKEMB-100 & 20{,}000 \\
 & IIITD-CLI & VistaFA2E / CIS 202 & 2{,}134 \\
 & LivDet13 (Clarkson) & Dalsa & 516 \\
 & LivDet15 (Clarkson) & LG IrisAccess EOU2200 / Dalsa & 1{,}906 \\
 & LivDet17 (Clarkson) &  LG IrisAccess EOU2200 & 3{,}954 \\
 & LivDet17 (Warsaw) & IrisGuard AD100 & 1{,}844 \\
 & NDCLD13 & LG4000 / IrisGuard AD100  & 1{,}700 \\
 & NDCLD15 & LG4000 / IrisGuard AD100  & 2{,}475 \\
 & Warsaw-BioBase-Disease & IrisGuard AD100  & 282 \\
 & Synthetic Iris & LG2200  & 4{,}129\\
 
\midrule
\multicolumn{3}{r}{\textbf{Subtotal (Bona fide)}} & \textbf{38{,}940} \\
\midrule

\multirow{2}{*}{\textbf{Non-intentional (Diseased)}}
 & Disease Eyes & MorphoTrust Mobile-Eyes  & 252 \\
 & Warsaw-BioBase-Disease & IrisGuard AD100  & 1{,}510 \\
\midrule
\multicolumn{3}{r}{\textbf{Subtotal (Diseased)}} & \textbf{1{,}762} \\
\midrule

\multirow{6}{*}{\textbf{Presentation Attacks (Textured Lenses)}}
 & IIITD-CLI & Vista Imaging FA2 / CIS 202 & 4{,}241 \\
 & LivDet13 (Clarkson) & Dalsa & 840 \\
 & LivDet15 (Clarkson) & LG IrisAccess EOU2200 / Dalsa & 2{,}547 \\
 & LivDet17 (Clarkson) & LG IrisAccess EOU2200 & 1{,}887 \\
 & NDCLD13 & LG4000 / IrisGuard AD100   & 3{,}400 \\
 & NDCLD15 & LG4000 / IrisGuard AD100  & 4{,}825 \\
\midrule
\multicolumn{3}{r}{\textbf{Subtotal (Textured Lenses)}} & \textbf{17{,}740} \\
\midrule

\multirow{3}{*}{\textbf{Presentation Attacks (Paper Prints)}}
 & LivDet15 (Clarkson) & LG IrisAccess EOU2200 / Dalsa & 3{,}492 \\
 & LivDet17 (Clarkson) & LG IrisAccess EOU2200 & 2{,}254 \\
 & LivDet17 (Warsaw) & IrisGuard AD100  & 2{,}669 \\
\midrule
\multicolumn{3}{r}{\textbf{Subtotal (Paper Prints)}} & \textbf{8{,}415} \\
\midrule

\multirow{2}{*}{\textbf{Synthetic}}
 & CASIA-IrisV4 & Learning-based synthesis & 10{,}000 \\
 & Synthetic Iris & Learning-based synthesis & 4{,}167 \\
\midrule
\multicolumn{3}{r}{\textbf{Subtotal (Synthetic)}} & \textbf{14{,}167} \\
\midrule

\textbf{Total} & & & \textbf{81{,}024} \\
\bottomrule
\end{tabular}
\end{table*}

\begin{table}[t]
\centering
\caption{VIS periocular image corpus (VSIA), used for cross-spectral evaluation (Protocol~3). Taxonomy follows the intentional/non-intentional/synthetic scheme used for the NIR datasets.}
\label{tab:vsia-summary}
\begin{tabular}{l l l}
\toprule
\textbf{PAI Taxonomy} & \textbf{Attack Subtype} & \textbf{\# Samples} \\
\midrule

\multirow{1}{*}{\textbf{Bona fide}}
 & --- & 5{,}200 \\
\midrule
\multicolumn{2}{r}{\textbf{Subtotal (Bona fide)}} & \textbf{5{,}200} \\
\midrule

\multirow{3}{*}{\textbf{Presentation Attacks}}
 & Paper prints & 5{,}200 \\
 & Electronic display & 10{,}400 \\
 & Print + display & 10{,}400 \\
\midrule
\multicolumn{2}{r}{\textbf{Subtotal (Presentation Attacks)}} & \textbf{26{,}000} \\
\midrule

\textbf{Total} & & \textbf{31{,}200} \\
\bottomrule
\end{tabular}
\end{table}

\begin{table*}[t]
\centering
\caption{Schematic overview of the three open-set evaluation protocols. ``Unseen'' denotes data not observed during training. Each held-out unit defines one experimental run.}
\label{tab:protocol-summary}
\begin{tabular}{l l l l}
\toprule
\textbf{Protocol} & \textbf{Training Data} & \textbf{Testing Data (Held-Out Unit)} & \textbf{Unseen Factor} \\
\midrule

\multirow{4}{*}{Protocol~1 (Cross-PAI)}
 & \multirow{4}{*}{NIR (all PAIs except held-out)}
 & Synthetic & \multirow{4}{*}{PAI type} \\
 & & Textured lenses & \\
 & & Paper prints & \\
 & & Diseased eyes & \\
\midrule

\multirow{10}{*}{Protocol~2 (Cross-Dataset)}
 & \multirow{10}{*}{NIR (all datasets except held-out)}
 & CASIA-IrisV4 & \multirow{10}{*}{Dataset / sensor} \\
 & & IIITD-CLI & \\
 & & LivDet13 (Clarkson) & \\
 & & LivDet15 (Clarkson) & \\
 & & LivDet17 (Clarkson) & \\
 & & LivDet17 (Warsaw) & \\
 & & NDCLD13 & \\
 & & NDCLD15 & \\
 & & Warsaw-BioBase-Disease & \\
 & & Synthetic Iris & \\
\midrule

\multirow{1}{*}{Protocol~3 (Cross-Spectral)}
 & \multirow{1}{*}{NIR (all datasets)}
 & VIS (VSIA) & \multirow{1}{*}{Spectral domain} \\
\bottomrule
\end{tabular}
\end{table*}

For reproducibility, we note that the three protocols yield a total of 15 distinct open-set evaluation runs. Protocol~1 generates four runs corresponding to the held-out PAIs {Synthetic, Textured lenses, Paper prints, Diseased eyes}. Protocol~2 generates ten runs, each corresponding to the exclusion of each of the following datasets: {CASIA-IrisV4, IIITD-CLI, LivDet13 (Clarkson), LivDet15 (Clarkson), LivDet17 (Clarkson), LivDet17 (Warsaw), NDCLD13, NDCLD15, Warsaw-BioBase-Disease, Synthetic Iris}. Protocol~3 yields a single run in which all NIR datasets are used for training and the VIS dataset (VSIA) is used exclusively for testing. No protocol-specific hyper-parameter tuning or retraining is performed; the same training configuration is applied across all runs.
\section{Results and Discussion}
\label{sec:results}

This section reports the empirical behaviour of the evaluated foundation models under the three open-set protocols and addresses the research questions stated in Section~1. Following the terminology of biometric Presentation Attack Detection (PAD), performance is reported using the Attack Presentation Classification Error Rate (APCER), Bona Fide Presentation Classification Error Rate (BPCER), and Detection Equal Error Rate (D-EER(\%)) \cite{ISO/IEC2015a}. APCER denotes the proportion of attack presentations incorrectly classified as bona fide, and therefore measures the vulnerability of the PAD system to spoof or attack presentations. BPCER denotes the proportion of bona fide presentations incorrectly classified as attacks, and therefore reflects the inconvenience imposed on genuine users. Lower values are better for both metrics.

D-EER(\%) is the operating point at which APCER and BPCER are equal, and provides a single summary measure of PAD separability. In addition to D-EER(\%), we report BPCER at fixed APCER operating points of 5\% and 10\% \cite{ISO/IEC2015a}. These operating points are important for PAD evaluation because practical systems often constrain the acceptable attack acceptance rate and then measure the corresponding rejection rate of bona fide users. Unless stated otherwise, all results are reported as D-EER(\%), BPCER@APCER~=~5\%, and BPCER@APCER~=~10\%, together with 95\% confidence intervals. Protocol~1 and Protocol~3 results are shown in Table~\ref{tab:prot13}, and Protocol~2 results in Table~\ref{tab:prot2}.

The results are organised according to the research questions in Section~1. Protocol~1 is used to answer RQ1, Protocol~2 addresses RQ2, Protocol~3 addresses RQ3, and the comparison between frozen  and LoRA-adapted models addresses RQ4. This organisation separates the empirical findings by source of distribution shift while using the same metrics and confidence interval procedure throughout.

\subsection{Confidence Interval Estimation and Interpretation}

To quantify sampling variability in the reported error rates, 95\% confidence intervals are estimated using a non-parametric sample-level bootstrap applied to the bona fide and attack score distributions. For each test condition, bona fide and attack scores are resampled independently with replacement while preserving class proportions, and the D-EER(\%) and BPCER@APCER metrics are recomputed for each bootstrap trial. This procedure yields the empirical distribution of each metric, from which percentile intervals are reported.

It is important to clarify what these confidence intervals represent. First, the bootstrap quantifies uncertainty arising from finite test sets rather than variability due to model training or hyper-parameter selection; the backbone parameters, LoRA configuration, and optimisation schedule are held fixed across all protocols. Second, because the same trained model is evaluated across all bootstrap samples, the reported intervals describe the variability of a deterministic decision function under new draws from the same test distribution. Third, no re-training or hyper-parameter re-tuning is performed across protocols, which is essential for a fair assessment of open-set behaviour. Consequently, the confidence intervals should be interpreted as measures of statistical stability in the test-phase error estimates rather than as measures of algorithmic variance.

Across all three protocols, confidence intervals are narrow for most conditions, reflecting the large number of test samples and the use of deterministic evaluation without threshold tuning on the test split. Narrow intervals do not imply good performance; rather, they indicate that poor performance (e.g., high D-EER(\%)) is statistically stable and not attributable to test-set randomness.

\subsection{RQ1: Generalisation to Unseen Presentation Attack Instruments}

Protocol~1 evaluates transfer to unseen PAIs. Table~\ref{tab:prot13} shows consistently high D-EER(\%) across synthetic irises, textured contact lenses, paper prints and diseased eyes, with several conditions approaching chance performance. For textured lenses, D-EER(\%) typically approach   $50\%$, with BPCER@APCER$=5\%$ frequently exceeding $84$\%. Synthetic irises and paper prints show comparable degradation, and diseased eyes present lower D-EER(\%) for some models but remain challenging at strict APCER operating points. Confidence intervals are narrow for all PAIs, confirming that these observations are statistically stable.

These results indicate that the evaluated foundation models do not reliably capture PAI-specific cues that transfer to previously unseen instruments. From a PAD standpoint, this is consistent with the underlying invariance properties of vision transformers trained for semantic recognition; suppressing photometric and micro-texture variation is advantageous for object classification but detrimental when the discriminative signal lies in subtle print/display artefacts or biological irregularities.

Task adaptation via LoRA is not beneficial in Protocol~1. LoRA frequently increases D-EER(\%) and BPCER, e.g., CLIP ($44.55 \rightarrow 50.00$ D-EER (\%) on synthetic irises) and DINOv3 ($34.82 \rightarrow 49.74$), illustrating that adaptation improves fit to observed PAIs while harming robustness to unseen PAIs. Hence, \textbf{RQ1 is answered negatively}: current foundation-model representations do not generalise reliably to unseen PAIs, and naive adaptation may amplify the failure.

\subsection{RQ2: Robustness to Dataset-Level Shifts}

Protocol~2 evaluates dataset-level robustness under fixed sensing modality. In contrast to Protocol~1, substantially lower D-EER(\%) values are observed for several held-out datasets. With frozen backbones, DINOv2 and OpenVision achieve low D-EER(\%) on LivDet17 Warsaw ($7.27$ and $3.10$), indicating meaningful transfer when acquisition conditions overlap with the training corpora. Confidence intervals confirm that these improvements are not artefacts of sampling variability.

LoRA frequently produces large gains in this protocol. DINOv2 reduces D-EER(\%) to $4.81$ on LivDet17 Clarkson and $1.79$ on LivDet17 Warsaw, and DINOv3 achieves single-digit D-EER(\%) for multiple datasets. These results suggest that parameter-efficient adaptation can align pretrained representations to dataset-specific texture and sensor statistics relevant for PAD, but does not address the failure modes observed in Protocol~1.

However, the same LoRA configuration degrades performance markedly on other held-out datasets (e.g., CASIA and Synthetic Iris). This pattern indicates that the benefits of LoRA are conditional on the compatibility between training and test distributions. The Synthetic Iris dataset remains difficult for all models (typically $>47$ D-EER(\%)), suggesting that generative visual artefacts constitute a shift that is not bridged by transfer learning alone.

Overall, \textbf{RQ2 is answered with qualification}: dataset-level transfer is feasible for certain datasets and models, but robustness is uneven and sensitive to adaptation.

\subsection{RQ3: Transfer Across Imaging Spectra}

Protocol~3 evaluates NIR-to-VIS transfer. A pronounced performance collapse is observed for most models, with high D-EER(\%) and very high BPCER at strict APCER. OpenVision without LoRA exhibits partial transfer ($18.12$ D-EER(\%)), but the advantage does not hold at operating points relevant for PAD, where BPCER approaches saturation for APCER$=5\%$. LoRA removes the advantage ($47.71$ D-EER(\%)) and drives BPCER close to $100$\%. Confidence intervals again confirm the stability of these effects. Hence, \textbf{RQ3 is answered negatively}: spectrum transfer remains a major failure mode for iris PAD, and parameter-efficient adaptation does not resolve the discrepancy.

\subsection{RQ4: Does LoRA Mitigate or Amplify Open-Set Failure?}

Across protocols, LoRA exhibits a clear trade-off. In Protocol~2, LoRA frequently reduces D-EER(\%) to single-digit values, indicating beneficial alignment under fixed modality. In Protocol~1 and Protocol~3, LoRA consistently amplifies failure, particularly for synthetic PAIs, paper prints and VIS transfer. These patterns indicate that LoRA increases sensitivity to the training distribution: beneficial when shifts are within-distribution (dataset-level), detrimental when shifts arise from unseen attack mechanisms or spectral changes. Confidence intervals show that these effects are statistically stable rather than random fluctuations. Therefore, LoRA should not be interpreted as a general-purpose solution for open-set PAD; its effect depends on the nature of the underlying shift.

\begin{table*}[t]
\centering
\caption{Open-set PAD performance under Protocol~1 and Protocol~3. Protocol~1 evaluates generalisation to unseen PAIs in NIR data, while Protocol~3 evaluates NIR$\rightarrow$VIS cross-spectral transfer using VSIA. All values are percentages and are reported as mean $\pm$ 95\% confidence interval. BPCER@5 and BPCER@10 denote BPCER measured at APCER fixed to 5\% and 10\%, respectively. Lower values indicate better performance, and the best result for each metric and test condition is shown in bold.}
\label{tab:prot13}
\scalebox{0.79}{
\begin{tabular}{
l c
*{4}{c}
c
}
\toprule
\multirow{2}{*}{\textbf{Model}} &
\multirow{2}{*}{\textbf{Metric}} &
\multicolumn{4}{c}{\textbf{Protocol~1 (Cross-PAI, NIR$\rightarrow$NIR)}} &
\textbf{Protocol~3 (Cross-Spectral, NIR$\rightarrow$VIS)} \\
\cmidrule(lr){3-6}\cmidrule(lr){7-7}
& & Synthetic & Textured & Diseased & Paper Prints & VSIA \\
\midrule

\multirow{3}{*}{CLIP (frozen)}
& D-EER (\%) $\downarrow$ & 44.55$\pm$0.58 & 50.00$\pm$0.56 & 32.66$\pm$2.89 & 50.00$\pm$0.76 & 39.23$\pm$0.98 \\
& BPCER5 (\%) $\downarrow$ & 98.24$\pm$0.23 & 84.74$\pm$0.61 & 84.75$\pm$4.44 & 99.30$\pm$0.20 & 76.35$\pm$1.28 \\
& BPCER10 (\%) $\downarrow$ & 95.59$\pm$0.39 & 84.74$\pm$0.61 & 72.70$\pm$6.38 & 98.03$\pm$0.47 & 67.88$\pm$1.41 \\
\midrule

\multirow{3}{*}{CLIP (LoRA)}
& D-EER (\%) $\downarrow$ & 50.00$\pm$0.49 & 50.00$\pm$0.55 & 36.16$\pm$3.40 & 50.00$\pm$0.79 & 50.00$\pm$0.71 \\
& BPCER5 (\%) $\downarrow$ & 98.65$\pm$0.20 & 90.43$\pm$0.50 & 82.98$\pm$5.32 & 96.50$\pm$0.60 & 92.50$\pm$0.89 \\
& BPCER10 (\%) $\downarrow$ & 96.74$\pm$0.30 & 90.43$\pm$0.50 & 69.86$\pm$5.85 & 89.80$\pm$1.03 & 85.38$\pm$1.06 \\
\midrule

\multirow{3}{*}{EVA02 (frozen)}
& D-EER (\%) $\downarrow$ & 41.72$\pm$0.50 & 50.00$\pm$0.55 & 47.54$\pm$3.17 & 48.09$\pm$0.77 & 50.00$\pm$0.86 \\
& BPCER5 (\%) $\downarrow$ & 84.49$\pm$0.78 & 95.51$\pm$0.48 & 94.33$\pm$3.36 & 92.37$\pm$1.22 & 98.65$\pm$0.31 \\
& BPCER10  (\%) $\downarrow$& 76.65$\pm$0.93 & 90.97$\pm$0.59 & 84.04$\pm$5.50 & \best{81.22\pm1.33} & 94.81$\pm$0.81 \\
\midrule

\multirow{3}{*}{EVA02 (LoRA)}
& D-EER (\%) $\downarrow$ & 43.32$\pm$0.53 & 50.28$\pm$0.57 & 39.37$\pm$2.91 & \best{44.61\pm0.80} & 46.58$\pm$0.70 \\
& BPCER5 (\%) $\downarrow$ & 87.97$\pm$0.74 & 90.98$\pm$0.51 & 90.07$\pm$3.91 & 94.95$\pm$0.81 & 88.85$\pm$0.88 \\
& BPCER10 (\%) $\downarrow$ & 80.51$\pm$0.77 & 90.98$\pm$1.37 & 83.69$\pm$5.14 & 88.14$\pm$1.12 & 84.42$\pm$1.00 \\
\midrule

\multirow{3}{*}{DINOv2 (frozen)}
& D-EER (\%) $\downarrow$ & \best{32.47\pm0.46} & 50.00$\pm$0.58 & 38.97$\pm$2.61 & 41.15$\pm$0.79 & 45.58$\pm$0.76 \\
& BPCER5 (\%) $\downarrow$ & \best{66.74\pm1.03} & 95.10$\pm$0.52 & 95.39$\pm$3.37 & 96.07$\pm$0.73 & 92.12$\pm$1.05 \\
& BPCER10 (\%) $\downarrow$ & \best{58.38\pm0.91} & 91.07$\pm$0.65 & 85.11$\pm$4.61 & 88.46$\pm$1.23 & 86.92$\pm$1.13 \\
\midrule

\multirow{3}{*}{DINOv2 (LoRA)}
& D-EER (\%) $\downarrow$ & 42.83$\pm$0.50 & 50.00$\pm$0.58 & \best{31.21\pm3.13} & 50.00$\pm$0.75 & 47.13$\pm$0.74 \\
& BPCER5 (\%) $\downarrow$ & 87.16$\pm$0.90 & 94.28$\pm$0.39 & \best{79.43\pm5.50} & 99.99$\pm$0.03 & 94.04$\pm$1.04 \\
& BPCER10 (\%) $\downarrow$ & 79.15$\pm$0.88 & 90.33$\pm$0.80 & \best{68.79\pm6.92} & 99.75$\pm$0.16 & 87.12$\pm$1.07 \\
\midrule

\multirow{3}{*}{DINOv3 (frozen)}
& D-EER (\%) $\downarrow$ & 34.82$\pm$0.49 & 50.00$\pm$0.59 & 50.00$\pm$3.23 & 51.68$\pm$0.79 & 42.27$\pm$0.66 \\
& BPCER5 (\%) $\downarrow$ & 73.51$\pm$1.07 & 94.31$\pm$0.42 & 95.39$\pm$2.66 & 99.08$\pm$0.25 & 95.96$\pm$0.91 \\
& BPCER10 (\%) $\downarrow$ & 63.77$\pm$0.98 & 90.71$\pm$0.60 & 90.07$\pm$3.73 & 97.92$\pm$0.37 & 88.85$\pm$1.23 \\
\midrule

\multirow{3}{*}{DINOv3 (LoRA)}
& D-EER (\%) $\downarrow$ & 49.74$\pm$0.55 & 50.00$\pm$0.57 & 39.26$\pm$3.16 & 50.00$\pm$0.76 & 44.04$\pm$0.83 \\
& BPCER5 (\%) $\downarrow$ & 96.90$\pm$0.38 & \best{84.44\pm0.61} & 86.17$\pm$5.49 & \best{86.37\pm0.82} & 93.46$\pm$0.95 \\
& BPCER10 (\%) $\downarrow$ & 92.73$\pm$0.56 & \best{84.44\pm0.61} & 83.69$\pm$4.96 & 82.40$\pm$0.90 & 83.27$\pm$1.29 \\
\midrule

\multirow{3}{*}{OpenVision (frozen)}
& D-EER (\%) $\downarrow$ & 41.42$\pm$0.52 & 50.00$\pm$0.57 & 43.25$\pm$2.88 & 44.73$\pm$0.80 & \best{18.12\pm0.57} \\
& BPCER5  (\%) $\downarrow$& 89.42$\pm$0.77 & 94.45$\pm$0.39 & 95.04$\pm$2.84 & 99.55$\pm$0.16 & \best{45.77\pm1.62} \\
& BPCER10 (\%) $\downarrow$ & 81.55$\pm$1.02 & 91.17$\pm$0.84 & 87.59$\pm$4.25 & 98.39$\pm$0.39 & \best{31.92\pm1.74} \\
\midrule

\multirow{3}{*}{OpenVision (LoRA)}
& D-EER (\%) $\downarrow$ & 50.00$\pm$0.53 & 50.00$\pm$0.59 & 39.72$\pm$3.59 & 50.00$\pm$0.75 & 47.71$\pm$0.60 \\
& BPCER5 (\%) $\downarrow$ & 94.04$\pm$0.56 & 91.81$\pm$0.48 & 87.59$\pm$4.44 & 100.00$\pm$0.02 & 99.62$\pm$0.22 \\
& BPCER10  (\%) $\downarrow$ & 88.61$\pm$0.66 & 91.81$\pm$0.48 & 80.85$\pm$5.67 & 99.65$\pm$0.13 & 97.12$\pm$0.54 \\
\bottomrule
\end{tabular}
}
\end{table*}

\begin{table*}[t]
\centering
\caption{Open-set PAD performance under Protocol~2 (cross-dataset NIR$\rightarrow$NIR evaluation). For each run, one complete dataset is held out for testing and the remaining NIR datasets are used for training and validation. All values are percentages and are reported as mean $\pm$ 95\% confidence interval. BPCER@5 and BPCER@10 denote BPCER measured at APCER fixed to 5\% and 10\%, respectively. Lower values indicate better performance, and the best result for each metric and held-out dataset is shown in bold.}
\label{tab:prot2}
\footnotesize
\setlength{\tabcolsep}{2.6pt}
\renewcommand{\arraystretch}{1.08}
\resizebox{\textwidth}{!}{
\begin{tabular}{l c *{10}{c}}
\toprule
\multirow{2}{*}{\textbf{Model}} &
\multirow{2}{*}{\textbf{Metric}} &
\multicolumn{10}{c}{\textbf{Held-out Test Dataset (Protocol~2)}} \\
\cmidrule(lr){3-12}
& &
\makecell{\textbf{CASIA}\\\textbf{IrisV4}} &
\makecell{\textbf{IIITD}\\\textbf{CLI}} &
\makecell{\textbf{LivDet13}\\\textbf{Clarkson}} &
\makecell{\textbf{LivDet15}\\\textbf{Clarkson}} &
\makecell{\textbf{LivDet17}\\\textbf{Clarkson}} &
\makecell{\textbf{LivDet17}\\\textbf{Warsaw}} &
\makecell{\textbf{NDCLD}\\\textbf{13}} &
\makecell{\textbf{NDCLD}\\\textbf{15}} &
\makecell{\textbf{Warsaw}\\\textbf{Disease}} &
\makecell{\textbf{Synthetic}\\\textbf{Iris}} \\
\midrule

\multirow{3}{*}{\makecell{CLIP \\ (frozen)}} 
& D-EER (\%) $\downarrow$ & $20.26 \pm 0.48$ & $33.88 \pm 1.21$ & $28.31 \pm 2.32$ & $43.91 \pm 1.14$ & $21.70 \pm 0.86$ & $12.59 \pm 0.92$ & $24.71 \pm 1.19$ & $21.66 \pm 1.02$ & $34.07 \pm 3.00$ & $50.00 \pm 1.11$ \\
& BPCER5 (\%) $\downarrow$ & $54.78 \pm 1.93$ & $87.21 \pm 2.02$ & $73.64 \pm 6.69$ & $98.48 \pm 0.60$ & $56.58 \pm 2.45$ & $30.53 \pm 4.37$ & $66.06 \pm 3.64$ & $61.62 \pm 3.66$ & $91.49 \pm 3.73$ & $94.79 \pm 0.92$ \\
& BPCER10 (\%) $\downarrow$ & $38.84 \pm 1.60$ & $76.05 \pm 2.55$ & $61.05 \pm 7.37$ & $95.96 \pm 1.00$ & $42.62 \pm 2.84$ & $15.89 \pm 2.22$ & $51.24 \pm 4.45$ & $44.85 \pm 3.94$ & $78.37 \pm 5.85$ & $91.06 \pm 1.03$ \\

\midrule

\multirow{3}{*}{\makecell{CLIP \\ (LoRA)}}
& D-EER (\%) $\downarrow$ & $30.34 \pm 0.49$ & $26.90 \pm 1.10$ & $35.31 \pm 2.67$ & $23.50 \pm 1.09$ & $11.88 \pm 0.75$ & $21.91 \pm 1.16$ & $22.90 \pm 1.18$ & $18.31 \pm 0.89$ & $33.70 \pm 2.69$ & $50.00 \pm 1.04$ \\
& BPCER5 (\%) $\downarrow$ & $61.78 \pm 1.32$ & $76.57 \pm 3.40$ & $78.88 \pm 5.14$ & $65.06 \pm 3.41$ & $33.59 \pm 4.92$ & $98.32 \pm 3.36$ & $61.12 \pm 3.65$ & $48.20 \pm 3.18$ & $80.50 \pm 4.96$ & $95.71 \pm 0.73$ \\
& BPCER10 (\%) $\downarrow$ & $50.87 \pm 1.21$ & $62.09 \pm 3.66$ & $68.22 \pm 4.85$ & $45.86 \pm 2.80$ & $15.02 \pm 2.12$ & $61.50 \pm 8.38$ & $42.41 \pm 3.27$ & $33.33 \pm 3.03$ & $69.15 \pm 6.57$ & $92.37 \pm 0.95$ \\

\midrule

\multirow{3}{*}{\makecell{EVA02 \\ (frozen)}} 
& D-EER (\%) $\downarrow$ & $26.82 \pm 0.52$ & $36.41 \pm 1.25$ & $36.65 \pm 2.70$ & $36.92 \pm 1.16$ & $31.71 \pm 1.05$ & $7.87 \pm 0.76$ & $30.96 \pm 1.27$ & $35.55 \pm 1.14$ & $46.41 \pm 2.88$ & $50.00 \pm 1.07$ \\
& BPCER5 (\%) $\downarrow$ & $58.28 \pm 1.54$ & $92.78 \pm 1.64$ & $89.34 \pm 4.26$ & $98.06 \pm 0.90$ & $78.15 \pm 2.36$ & $13.83 \pm 3.04$ & $80.12 \pm 3.21$ & $85.13 \pm 2.20$ & $93.62 \pm 3.37$ & $95.33 \pm 0.90$ \\
& BPCER10 (\%) $\downarrow$ & $46.52 \pm 1.27$ & $83.27 \pm 2.46$ & $76.55 \pm 5.91$ & $92.18 \pm 1.92$ & $68.16 \pm 1.94$ & $5.59 \pm 1.27$ & $67.88 \pm 3.17$ & $75.56 \pm 2.20$ & $85.46 \pm 4.96$ & $91.28 \pm 1.04$ \\

\midrule

\multirow{3}{*}{\makecell{EVA02 \\ (LoRA)}} 
& D-EER (\%) $\downarrow$ & $39.06 \pm 0.59$ & $36.78 \pm 1.18$ & $33.77 \pm 2.50$ & $37.72 \pm 1.11$ & $24.20 \pm 0.96$ & $13.50 \pm 1.01$ & $28.53 \pm 1.30$ & $33.86 \pm 1.11$ & $44.69 \pm 3.42$ & $50.00 \pm 1.11$ \\
& BPCER5 (\%) $\downarrow$ & $75.30 \pm 1.09$ & $90.58 \pm 1.45$ & $91.09 \pm 4.16$ & $96.69 \pm 0.95$ & $66.79 \pm 2.69$ & $39.32 \pm 4.83$ & $77.88 \pm 3.35$ & $84.57 \pm 2.19$ & $93.26 \pm 3.01$ & $95.93 \pm 0.70$ \\
& BPCER10  (\%) $\downarrow$ & $66.14 \pm 1.20$ & $82.99 \pm 1.99$ & $80.62 \pm 5.04$ & $93.34 \pm 1.20$ & $51.64 \pm 3.37$ & $20.66 \pm 3.11$ & $61.71 \pm 3.32$ & $73.66 \pm 2.62$ & $87.23 \pm 4.44$ & $91.35 \pm 1.14$ \\

\midrule

\multirow{3}{*}{\makecell{DINOv2 \\ (frozen)}} 
& D-EER (\%) $\downarrow$ & \boldmath{$11.78 \pm 0.38$} & $21.08 \pm 1.04$ & $23.29 \pm 2.21$ & $19.21 \pm 0.93$ & $18.01 \pm 0.83$ & $7.27 \pm 0.79$ & $17.24 \pm 1.06$ & $15.08 \pm 0.86$ & $40.79 \pm 2.88$ & $50.00 \pm 1.11$ \\
& BPCER5 (\%) $\downarrow$ & \boldmath{$24.51 \pm 1.85$} & $67.39 \pm 3.59$ & $51.74 \pm 7.08$ & $74.40 \pm 4.33$ & $52.33 \pm 3.85$ & $11.06 \pm 2.77$ & $47.18 \pm 3.74$ & $41.21 \pm 3.54$ & $89.72 \pm 4.08$ & $95.33 \pm 0.77$ \\
& BPCER10 (\%) $\downarrow$ & \boldmath{$13.98 \pm 0.96$} & $46.34 \pm 3.91$ & $41.47 \pm 5.62$ & $43.97 \pm 3.89$ & $34.72 \pm 3.34$ & $4.39 \pm 1.38$ & $29.71 \pm 3.29$ & $23.76 \pm 2.50$ & $81.91 \pm 4.97$ & $91.21 \pm 1.17$ \\

\midrule

\multirow{3}{*}{\makecell{DINOv2 \\ (LoRA)}} 
& D-EER (\%) $\downarrow$ & $32.61 \pm 0.55$ & \boldmath{$10.68 \pm 0.77$} & $22.27 \pm 2.21$ & $25.87 \pm 1.09$ & $4.81 \pm 0.53$ & \boldmath{$1.79 \pm 0.41$} & \boldmath{$3.53 \pm 0.55$} & \boldmath{$6.95 \pm 0.65$} & $33.32 \pm 2.85$ & $50.00 \pm 1.07$ \\
& BPCER5 (\%) $\downarrow$ & $73.17 \pm 1.55$ & \boldmath{$31.16 \pm 5.13$} & $44.19 \pm 5.82$ & $84.21 \pm 2.50$ & $4.40 \pm 1.70$ & \boldmath{$0.00 \pm 0.06$} & \boldmath{$2.41 \pm 0.85$} & \boldmath{$10.83 \pm 1.88$} & $76.60 \pm 6.74$ & $96.85 \pm 0.69$ \\
& BPCER10 (\%) $\downarrow$ & $60.29 \pm 1.41$ & \boldmath{$11.43 \pm 2.16$} & $34.88 \pm 5.23$ & $67.31 \pm 2.89$ & $1.04 \pm 0.38$ & \boldmath{$0.00 \pm 0.00$} & \boldmath{$1.47 \pm 0.53$} & \boldmath{$4.93 \pm 0.95$} & $65.96 \pm 6.03$ & $93.27 \pm 0.93$ \\

\midrule

\multirow{3}{*}{\makecell{DINOv3 \\ (frozen)}} 
& D-EER (\%) $\downarrow$ & $25.44 \pm 0.52$ & $27.61 \pm 1.05$ & $23.45 \pm 2.32$ & $21.67 \pm 0.97$ & $14.34 \pm 0.81$ & $7.53 \pm 0.78$ & $17.60 \pm 0.99$ & $18.07 \pm 0.96$ & $47.16 \pm 2.98$ & $50.00 \pm 1.06$ \\
& BPCER5 (\%) $\downarrow$ & $65.72 \pm 1.82$ & $81.54 \pm 3.05$ & $52.13 \pm 6.88$ & $73.24 \pm 3.65$ & $38.19 \pm 3.28$ & $12.58 \pm 2.96$ & $52.82 \pm 4.15$ & $49.33 \pm 3.77$ & $92.91 \pm 3.19$ & $97.48 \pm 0.58$ \\
& BPCER10 (\%) $\downarrow$ & $51.25 \pm 1.82$ & $66.17 \pm 3.07$ & $38.57 \pm 4.95$ & $50.16 \pm 3.91$ & $22.23 \pm 2.60$ & $4.12 \pm 1.44$ & $32.35 \pm 3.32$ & $32.85 \pm 2.93$ & $85.46 \pm 4.26$ & $94.82 \pm 0.81$ \\

\midrule

\multirow{3}{*}{\makecell{DINOv3 \\ (LoRA)}} 
& D-EER (\%) $\downarrow$ & $50.00 \pm 0.57$ & $14.48 \pm 0.92$ & \boldmath{$7.90 \pm 1.51$} & \boldmath{$10.13 \pm 0.66$} & \boldmath{$2.32 \pm 0.35$} & $7.64 \pm 0.92$ & $6.12 \pm 0.71$ & $8.33 \pm 0.67$ & \boldmath{$31.95 \pm 2.86$} & \boldmath{$47.66 \pm 1.04$} \\
& BPCER5 (\%) $\downarrow$ & $94.85 \pm 0.50$ & $35.94 \pm 3.63$ & \boldmath{$15.12 \pm 7.08$} & \boldmath{$31.27 \pm 4.83$} & \boldmath{$0.66 \pm 0.31$} & $8.68 \pm 1.30$ & $8.35 \pm 2.35$ & $12.93 \pm 1.96$ & \boldmath{$74.47 \pm 5.67$} & $94.07 \pm 0.72$ \\
& BPCER10 (\%) $\downarrow$ & $90.08 \pm 0.57$ & $20.71 \pm 2.55$ & \boldmath{$6.01 \pm 2.72$} & \boldmath{$10.28 \pm 2.54$} & \boldmath{$0.23 \pm 0.15$} & $6.94 \pm 1.16$ & $3.76 \pm 0.94$ & $6.83 \pm 1.27$ & \boldmath{$63.83 \pm 6.20$} & $90.39 \pm 1.77$ \\

\midrule

\multirow{3}{*}{\makecell{OpenVision \\ (frozen)}} 
& D-EER (\%) $\downarrow$ & $19.63 \pm 0.49$ & $31.31 \pm 1.11$ & $21.53 \pm 2.28$ & $27.91 \pm 0.98$ & $16.42 \pm 0.85$ & $3.10 \pm 0.55$ & $19.71 \pm 1.12$ & $25.37 \pm 1.01$ & $43.25 \pm 3.14$ & $49.80 \pm 1.10$ \\
& BPCER5 (\%) $\downarrow$ & $57.62 \pm 3.14$ & $81.82 \pm 2.72$ & $52.13 \pm 6.49$ & $92.08 \pm 1.76$ & $51.44 \pm 3.30$ & $0.38 \pm 0.52$ & $57.76 \pm 3.56$ & $77.17 \pm 2.52$ & $93.62 \pm 3.72$ & \boldmath{$93.75 \pm 0.86$} \\
& BPCER10 (\%) $\downarrow$ & $37.15 \pm 2.16$ & $69.40 \pm 2.84$ & $37.02 \pm 7.36$ & $78.23 \pm 2.39$ & $33.71 \pm 3.34$ & \boldmath{$0.00 \pm 0.00$} & $36.59 \pm 3.68$ & $58.99 \pm 3.23$ & $85.11 \pm 4.61$ & \boldmath{$89.42 \pm 1.16$} \\

\midrule

\multirow{3}{*}{\makecell{OpenVision \\ (LoRA)}} 
& D-EER (\%) $\downarrow$ & $50.00 \pm 0.51$ & $20.63 \pm 0.99$ & $12.24 \pm 1.82$ & $14.27 \pm 0.97$ & $9.44 \pm 0.67$ & $6.18 \pm 0.75$ & $22.46 \pm 1.31$ & $12.93 \pm 0.76$ & $39.07 \pm 2.76$ & $50.00 \pm 1.01$ \\
& BPCER5 (\%) $\downarrow$ & $99.23 \pm 0.13$ & $57.92 \pm 4.29$ & $33.72 \pm 12.30$ & $42.03 \pm 4.99$ & $14.77 \pm 1.76$ & $7.16 \pm 1.36$ & $41.41 \pm 3.68$ & $35.07 \pm 3.92$ & $95.04 \pm 3.73$ & $98.74 \pm 0.38$ \\
& BPCER10 (\%) $\downarrow$ & $98.37 \pm 0.21$ & $39.74 \pm 2.61$ & $19.19 \pm 7.17$ & $19.57 \pm 2.42$ & $9.13 \pm 1.17$ & $1.74 \pm 1.03$ & $31.82 \pm 2.41$ & $18.63 \pm 2.46$ & $85.11 \pm 4.62$ & $97.02 \pm 0.63$ \\
\bottomrule
\end{tabular}
}
\end{table*}
\subsection{Summary of Failure Modes}

Taken together, the results across the three protocols indicate that the dominant failure modes of the evaluated vision foundation models arise from shifts in \emph{attack mechanism} and \emph{sensing spectrum}, rather than from dataset-level variation alone. Cross-dataset transfer (Protocol~2) can be strong, particularly for self-supervised transformers and when limited adaptation via LoRA is applied, but this improvement does not extend to unseen PAIs (Protocol~1) or to cross-spectral operation (Protocol~3). In these latter regimes, D-EER(\%) remains high and BPCER at strict APCER operating points is often close to saturation, with narrow confidence intervals indicating that the degradation is statistically stable.

These findings support two main conclusions. First, strong performance under constrained or semi-closed evaluation settings must not be interpreted as evidence of robust open-set security. Second, open-set iris PAD requires representations that remain sensitive to presentation artefacts while being stable across sensors and spectra. The following analysis examines these failure modes at the feature level to better understand how foundation-model embeddings react to different types of shift.

Section~5 establishes where the models fail in terms of PAD operating metrics. Section~6 examines why these failures occur by analysing whether bona fide and attack samples remain geometrically separable in the learned feature space.
\section{Feature-Space Separability Analysis}
\label{sec:sep-analysis}

To better understand how open-set shifts affect the embedding geometry of foundation models, we perform a lightweight feature-space separability analysis. For each model and condition, let $\mu_{\mathrm{BF}}$ and $\mu_{\mathrm{AT}}$ denote the mean feature vectors of bona fide and attack samples, respectively, and let $\sigma_{\mathrm{BF}}$ and $\sigma_{\mathrm{AT}}$ denote the corresponding root mean squared distances from class samples to their class means (i.e., intra-class dispersion). We define the \emph{separability ratio} \cite{bauer1998linear}:
\[
R = \frac{\lVert \mu_{\mathrm{BF}} - \mu_{\mathrm{AT}} \rVert_2}{\sigma_{\mathrm{BF}} + \sigma_{\mathrm{AT}}},
\]
which captures how far apart the class means are relative to their combined spread. Higher $R$ indicates better linear separability in the embedding space.

To quantify the effect of a shift, we measure separability in-domain (validation distribution) and under shift (held-out PAI, dataset or spectrum), denoted $R_{\mathrm{in}}$ and $R_{\mathrm{shift}}$, respectively. We report the \emph{Separability Ratio Drop} (SRD) \cite{bauer1998linear},
\[
\mathrm{SRD} = \frac{R_{\mathrm{in}} - R_{\mathrm{shift}}}{R_{\mathrm{in}}} \times 100,
\]
which is positive when separability degrades under shift ($R_{\mathrm{shift}} < R_{\mathrm{in}}$) and negative when separability improves ($R_{\mathrm{shift}} > R_{\mathrm{in}}$).

To assess how shifts affect intra-class structure, we further compute the \emph{Dispersion Drop Percentage} (DDP) \cite{bauer1998linear} as the relative change in total dispersion,
\[
\mathrm{DDP} = \frac{(\sigma_{\mathrm{BF}} + \sigma_{\mathrm{AT}})_{\mathrm{in}} - (\sigma_{\mathrm{BF}} + \sigma_{\mathrm{AT}})_{\mathrm{shift}}}{(\sigma_{\mathrm{BF}} + \sigma_{\mathrm{AT}})_{\mathrm{in}}} \times 100.
\]
Positive DDP indicates a reduction in total dispersion (tighter class-specific clusters), whereas negative DDP indicates an expansion of the feature clouds. 

In the tables below, SRD and DDP should be interpreted as relative changes from the in-domain validation distribution to the shifted test distribution. A positive SRD indicates that bona fide and attack features become less separable under shift, whereas a negative SRD indicates that separability is preserved or improves. For DDP, positive values indicate that the total within-class dispersion decreases under shift, producing tighter class-specific feature clouds, whereas negative values indicate dispersion expansion. Thus, lower SRD is preferable, while higher DDP is preferable. SRD and DDP should be interpreted together: a low or negative SRD with positive DDP indicates favourable feature-space behaviour, whereas a high positive SRD and negative DDP indicates separability collapse with increased feature spread.
This analysis does not introduce new decision rules; instead, it provides a geometric view that complements the PAD metrics reported in Section~5.

\vspace{0.15cm}
\noindent\textbf{Protocol~1 (Unseen Presentation Attacks).}
Under cross-PAI shift, separability collapses substantially for LoRA-adapted models, with $\mathrm{SRD}$ typically in the range of $+50$ to $+90$ percentage points across synthetic irises, textured contact lenses, diseased eyes and paper prints (Table~\ref{tab:sep-prot13}). frozen feature extractors exhibit more mixed behaviour: some configurations preserve or even improve separability on particular PAIs (negative SRD), whereas others show marked degradation. The strong positive SRD observed for many LoRA configurations indicates pronounced over-specialisation to PAI-specific artefacts present during training, at the expense of open-set generalisation.

The DDP values provide additional insight. For several LoRA configurations, SRD is large and positive while DDP is negative or only weakly positive, indicating that collapse is driven not only by reduced inter-class distance but also by increased intra-class dispersion. In other words, bona fide and attack manifolds become more entangled under unseen PAIs, which is consistent with the high D-EER(\%) and saturated BPCER observed in Protocol~1.

\vspace{0.15cm}
\noindent\textbf{Protocol~2 (Cross-Dataset).}
Across cross-dataset shifts, foundation models largely preserve separability (many SRD values are close to or below zero), and in several cases separability improves on the held-out dataset. This is particularly pronounced for self-supervised transformers (DINOv2/DINOv3) and OpenVision, where SRD is frequently negative on LivDet13/15/17-Clarkson and NDCLD13/15 (Table~\ref{tab:sep-prot2}). The corresponding positive DDP values indicate a consistent reduction in intra-class dispersion, i.e., tighter bona fide and attack clusters relative to the in-domain reference.

LoRA adaptation further improves separability in some cross-dataset conditions, but its effect is uneven and depends on the specific held-out dataset. Synthetic Iris again stands out as an adverse case: SRD is strongly positive for many models, indicating a marked reduction in separability. This is consistent with the performance results in Protocol~2, where Synthetic Iris remains difficult despite transfer learning and adaptation, and suggests that its generative characteristics induce a mismatch that is not captured by the in-domain feature geometry.

\vspace{0.15cm}
\noindent\textbf{Protocol~3 (Cross-Spectral VIS--NIR).}
Cross-spectral evaluation constitutes the most severe open-set shift in the separability analysis. On VSIA, LoRA consistently amplifies separability collapse, with large positive SRD values across all investigated models except DINOv3 (Table~\ref{tab:sep-prot13}). Negative DDP values for several model–LoRA combinations further suggest that feature clouds expand under VIS imaging, leading to increased overlap between bona fide and attack distributions in the embedding space. This is consistent with the high D-EER(\%) and saturated BPCER observed in Protocol~3.

Interestingly, several frozen models display negative SRD in the cross-spectral setting, indicating that, in purely geometric terms, bona fide and attacks remain separable in $\mathbb{R}^d$ despite degraded operating points at the score level. This suggests that part of the cross-spectral failure arises at the decision layer (e.g., threshold misalignment between NIR and VIS) rather than from a complete collapse of the underlying feature geometry.

\vspace{0.15cm}
\noindent\textbf{Summary.}
The separability analysis reveals three distinct regimes. First, cross-dataset shifts tend to preserve or even improve separability, especially for self-supervised transformers, which aligns with the relatively strong D-EER(\%) observed in Protocol~2. Second, unseen PAIs induce substantial separability collapse, driven by a combination of reduced inter-class distance and increased intra-class dispersion, which mirrors the severe degradation in Protocol~1. Third, cross-spectral shifts are the most severe: LoRA tends to amplify collapse, while frozen models sometimes retain a degree of geometric separability that is not exploited by a single global operating threshold.

Overall, these observations support the interpretation that LoRA overfits to PAI and dataset-specific artefacts, thereby reducing open-set generalisation, while frozen foundation models preserve more generic structure but lack PAD-specific decision mechanisms. Importantly, the analysis also shows that open-set PAD failures cannot be fully attributed to changes in global inter-class distance; changes in intra-class manifold structure play a central role, providing complementary insight beyond conventional PAD performance metrics.

\begin{table*}[t]
\caption{Feature-space separability analysis under Protocol~2. SRD denotes separability ratio drop and DDP denotes dispersion drop percentage, both computed relative to the in-domain validation distribution. Lower SRD is better; negative SRD indicates preserved or improved separability under shift. Higher DDP is better; positive DDP indicates tighter intra-class feature clouds, while negative DDP indicates dispersion expansion. Best values for each metric and held-out dataset are shown in bold.}
\label{tab:sep-prot2}
\centering
\footnotesize
\setlength{\tabcolsep}{2.8pt}
\renewcommand{\arraystretch}{1.08}
\resizebox{\textwidth}{!}{
\begin{tabular}{l c *{10}{c}}
\toprule
\multirow{2}{*}{\textbf{Model}} &
\multirow{2}{*}{\textbf{Metric}} &
\multicolumn{10}{c}{\textbf{Held-out Test Dataset (Protocol~2)}} \\
\cmidrule(lr){3-12}
& &
\makecell{\textbf{CASIA}\\\textbf{IrisV4}} &
\makecell{\textbf{IIITD}\\\textbf{CLI}} &
\makecell{\textbf{LivDet13}\\\textbf{Clarkson}} &
\makecell{\textbf{LivDet15}\\\textbf{Clarkson}} &
\makecell{\textbf{LivDet17}\\\textbf{Clarkson}} &
\makecell{\textbf{LivDet17}\\\textbf{Warsaw}} &
\makecell{\textbf{NDCLD}\\\textbf{13}} &
\makecell{\textbf{NDCLD}\\\textbf{15}} &
\makecell{\textbf{Warsaw}\\\textbf{Disease}} &
\makecell{\textbf{Synthetic}\\\textbf{Iris}} \\
\midrule

\multirow{2}{*}{CLIP (frozen)}
& SRD $\downarrow$ & \best{-377.10} & $62.37$ & $-39.10$ & $-31.13$ & $-67.68$ & $-228.94$ & $-21.14$ & $47.93$ & $-22.48$ & $-64.88$ \\
& DDP $\uparrow$& $24.99$ & $22.35$ & $11.31$ & $6.53$ & $10.19$ & $23.14$ & $22.29$ & $18.95$ & $25.29$ & $28.69$ \\
\midrule

\multirow{2}{*}{CLIP (LoRA)}
& SRD $\downarrow$ & $17.08$ & \best{-20.95} & $64.47$ & $64.50$ & $16.96$ & $14.46$ & $-5.53$ & $33.53$ & $67.08$ & $78.13$ \\
& DDP $\uparrow$ & $22.18$ & $31.43$ & $17.04$ & $-4.67$ & $13.15$ & $16.74$ & $25.84$ & $8.69$ & $18.76$ & $27.81$ \\
\midrule

\multirow{2}{*}{EVA02 (frozen)}
& SRD $\downarrow$ & $-44.86$ & $54.59$ & $-45.70$ & \best{-88.29} & \best{-288.55} & \best{-439.69} & $-29.88$ & $10.31$ & \best{-26.19} & $-7.47$ \\
& DDP $\uparrow$ & $20.56$ & $24.75$ & $9.93$ & $-12.27$ & $12.81$ & $23.54$ & \best{39.93} & $39.43$ & \best{43.62} & \best{38.19} \\
\midrule

\multirow{2}{*}{EVA02 (LoRA)}
& SRD $\downarrow$ & $6.35$ & $57.62$ & $36.70$ & $47.29$ & $-16.68$ & $-81.34$ & $25.67$ & $35.83$ & $68.30$ & $73.64$ \\
& DDP $\uparrow$& \best{25.09} & $14.05$ & \best{17.98} & $11.71$ & \best{25.21} & \best{30.26} & $24.37$ & $25.62$ & $27.34$ & $22.82$ \\
\midrule

\multirow{2}{*}{DINOv2 (frozen)}
& SRD $\downarrow$ & $-145.22$ & $48.52$ & $-19.49$ & $-28.72$ & $-97.80$ & $-176.81$ & $18.50$ & $41.14$ & $1.47$ & $29.12$ \\
& DDP $\uparrow$ & $-8.96$ & $35.59$ & $-6.99$ & $4.95$ & $8.88$ & $15.95$ & $40.08$ & \best{39.85} & $34.96$ & $35.01$ \\
\midrule

\multirow{2}{*}{DINOv2 (LoRA)}
& SRD $\downarrow$ & $14.83$ & $2.48$ & $55.52$ & $31.00$ & $14.61$ & $-31.85$ & $-29.23$ & \best{-28.65} & $59.09$ & $87.50$ \\
& DDP $\uparrow$ & $14.21$ & $1.43$ & $2.37$ & $-1.86$ & $5.01$ & $18.80$ & $13.70$ & $8.30$ & $21.80$ & $18.42$ \\
\midrule

\multirow{2}{*}{DINOv3 (frozen)}
& SRD $\downarrow$ & $-156.22$ & $51.29$ & $-7.86$ & $-38.07$ & $-85.17$ & $-245.57$ & $19.11$ & $25.07$ & $-19.36$ & $-1.39$ \\
& DDP $\uparrow$ & $14.00$ & $24.43$ & $-7.65$ & $-7.46$ & $25.17$ & $20.44$ & $29.05$ & $28.41$ & $32.72$ & $25.22$ \\
\midrule

\multirow{2}{*}{DINOv3 (LoRA)}
& SRD $\downarrow$ & $-6.36$ & $-3.50$ & $-47.29$ & $9.88$ & $-19.79$ & $-47.32$ & \best{-35.77} & $-17.77$ & $62.30$ & $51.38$ \\
& DDP $\uparrow$ & $23.66$ & $9.51$ & $15.95$ & \best{11.95} & $16.65$ & $18.39$ & $15.45$ & $8.01$ & $16.81$ & $16.32$ \\
\midrule

\multirow{2}{*}{OpenVision (frozen)}
& SRD $\downarrow$ & $-368.02$ & $40.82$ & \best{-64.68} & $-17.57$ & $-101.64$ & $-288.87$ & $21.57$ & $46.61$ & $4.47$ & \best{-88.63} \\
& DDP $\uparrow$ & $25.08$ & \best{36.41} & $11.26$ & $11.09$ & $18.68$ & $23.96$ & $18.90$ & $11.75$ & $32.24$ & $34.13$ \\
\midrule

\multirow{2}{*}{OpenVision (LoRA)}
& SRD $\downarrow$ & $58.96$ & $56.52$ & $42.36$ & $48.87$ & $14.88$ & $10.84$ & $64.78$ & $46.11$ & $81.50$ & $88.12$ \\
& DDP $\uparrow$ & $14.60$ & $-37.86$ & $0.25$ & $-21.96$ & $-1.32$ & $10.01$ & $2.43$ & $-16.61$ & $7.21$ & $5.04$ \\
\bottomrule
\end{tabular}
}
\end{table*}

\begin{table*}[t]
\centering
\caption{Feature-space separability analysis under Protocol~1 and Protocol~3. SRD denotes separability ratio drop and DDP denotes dispersion drop percentage, both computed relative to the in-domain validation distribution. Lower SRD is better; negative SRD indicates preserved or improved separability under shift. Higher DDP is better; positive DDP indicates tighter intra-class structure, while negative DDP indicates dispersion expansion. Best values for each metric and test condition are shown in bold.}
\label{tab:sep-prot13}
\resizebox{0.88\textwidth}{!}{%
\begin{tabular}{l c *{5}{c}}
\toprule
\multirow{2}{*}{\textbf{Model}} &
\multirow{2}{*}{\textbf{Metric}} &
\multicolumn{4}{c}{\textbf{Protocol~1 (Cross-PAI, NIR$\rightarrow$NIR)}} &
\textbf{Protocol~3 (Cross-Spectral, NIR$\rightarrow$VIS)} \\
\cmidrule(lr){3-6}\cmidrule(lr){7-7}
& & Synthetic & Textured lenses & Diseased & Paper Prints & VSIA \\
\midrule

\multirow{2}{*}{CLIP (frozen)}
& SRD $\downarrow$ & \best{-192.14} & $50.30$ & $-27.23$ & $-36.70$ & $-95.60$ \\
& DDP $\uparrow$ & \best{17.79} & $-0.08$ & $21.48$ & $-8.98$ & $11.54$ \\
\midrule

\multirow{2}{*}{CLIP (LoRA)}
& SRD $\downarrow$ & $58.15$ & $80.58$ & $59.81$ & $69.41$ & $67.83$ \\
& DDP $\uparrow$ & $8.85$ & $-10.34$ & $33.44$ & \best{20.27} & $11.56$ \\
\midrule

\multirow{2}{*}{EVA02 (frozen)}
& SRD $\downarrow$ & $9.13$ & \best{37.48} & $-18.86$ & \best{-170.00} & $-80.23$ \\
& DDP $\uparrow$ & $13.68$ & $9.39$ & \best{42.95} & $-23.87$ & $22.02$ \\
\midrule

\multirow{2}{*}{EVA02 (LoRA)}
& SRD $\downarrow$ & $42.43$ & $81.16$ & $55.44$ & $17.34$ & $47.98$ \\
& DDP $\uparrow$ & $15.84$ & $-1.97$ & $31.97$ & $-2.07$ & \best{27.65} \\
\midrule

\multirow{2}{*}{DINOv2 (frozen)}
& SRD $\downarrow$ & $-77.31$ & $48.92$ & $3.63$ & $-71.41$ & $-38.31$ \\
& DDP $\uparrow$& $-7.25$ & \best{25.69} & $34.33$ & $-6.33$ & $-23.27$ \\
\midrule

\multirow{2}{*}{DINOv2 (LoRA)}
& SRD $\downarrow$ & $69.84$ & $90.64$ & $58.02$ & $55.17$ & $75.82$ \\
& DDP $\uparrow$ & $1.83$ & $-11.71$ & $22.26$ & $2.41$ & $9.59$ \\
\midrule

\multirow{2}{*}{DINOv3 (frozen)}
& SRD $\downarrow$ & $-77.23$ & $51.08$ & $-15.73$ & $-87.54$ & \best{-156.22} \\
& DDP $\uparrow$ & $7.20$ & $10.86$ & $29.89$ & $-2.58$ & $-14.91$ \\
\midrule

\multirow{2}{*}{DINOv3 (LoRA)}
& SRD $\downarrow$ & $34.92$ & $78.59$ & $54.03$ & $44.03$ & $-20.12$ \\
& DDP $\uparrow$ & $13.95$ & $-5.59$ & $15.60$ & $8.67$ & $12.61$ \\
\midrule

\multirow{2}{*}{OpenVision (frozen)}
& SRD $\downarrow$ & $-148.27$ & $58.10$ & \best{-9.41} & $-83.03$ & $-151.92$ \\
& DDP $\uparrow$ & $15.07$ & $5.86$ & $28.93$ & $-3.99$ & $10.80$ \\
\midrule

\multirow{2}{*}{OpenVision (LoRA)}
& SRD $\downarrow$ & $81.27$ & $93.12$ & $86.16$ & $81.00$ & $93.13$ \\
& DDP $\uparrow$ & $-6.14$ & $-132.87$ & $1.97$ & $-21.99$ & $-34.30$ \\
\bottomrule
\end{tabular}
}
\end{table*}

\section{Implications for PAD Design}

The empirical results and feature-space analysis have direct implications for the design of future iris PAD systems based on vision foundation models.

\subsection{Reconciling Invariance with Artefact Sensitivity}

The first and perhaps most fundamental observation is the tension between the invariances engineered into vision foundation models and the requirements of PAD. The evaluated models are explicitly optimised to suppress nuisance variation such as local texture noise, illumination changes and sensor-specific artefacts in order to stabilise semantic recognition. In contrast, iris PAD often relies on precisely those low-level deviations, including subtle print artefacts, contact lens patterns and spectrum-dependent reflectance changes.

The Protocol~1 and Protocol~3 results indicate that, in their current form, foundation models tend to discard or de-emphasise the very cues that distinguish bona fide from attacks under unseen PAIs and cross-spectral conditions. The separability analysis supports this: under these shifts, both the inter-class distance and intra-class structure deteriorate, particularly after LoRA adaptation. For PAD design, this suggests that directly reusing generic visual backbones without PAD-aware objectives is insufficient. Instead, PAD-specific pretraining or fine-tuning should explicitly preserve, or even amplify, those local statistics that are predictive of presentation artefacts.

\subsection{Beyond Dataset-Level Transfer: PAI and Spectrum-Aware Training}

The cross-dataset experiments (Protocol~2) show that, under fixed sensing modality, foundation models can achieve low D-EER(\%) on held-out datasets, especially when combined with moderate adaptation. This indicates that such models can learn sensor-agnostic representations to some extent, and that their failure under Protocol~1 and Protocol~3 is not due to a complete lack of capacity.

From a PAD design standpoint, this suggests that standard domain adaptation or domain generalisation techniques that target dataset or sensor variation are not sufficient. Instead, explicit \emph{PAI- and spectrum-aware} training strategies are needed, for example:
\begin{itemize}
    \item stratified sampling and loss weighting that emphasise rare or synthetic PAIs rather than mirroring the natural class distribution;
    \item augmentation schemes that mimic realistic print/display artefacts and spectral distortions, with constraints informed by actual PAI generation processes;
    \item multi-domain training where NIR and VIS samples are treated as distinct but related domains, with explicit regularisers that encourage retention of PAD-relevant structure across spectra.
\end{itemize}
The present results indicate that, without such targeted interventions, strong dataset-level transfer can coexist with poor robustness to unseen attack mechanisms and spectra.

\subsection{Use of Parameter-Efficient Adaptation}

LoRA is attractive for PAD because it offers a low-parameter way to specialise large backbones. However, across the three protocols, LoRA exhibits a clear pattern: it often improves dataset-level robustness (Protocol~2) while degrading robustness under unseen PAIs and spectra (Protocol~1 and Protocol~3). The separability analysis shows that these degradations are accompanied by large positive SRD and, in many cases, unfavourable DDP, indicating that LoRA encourages over-specialisation to training PAIs and acquisition conditions.

For PAD design, two practical implications follow. First, parameter-efficient adaptation should not be applied indiscriminately; instead, it should be evaluated under the specific open-set conditions relevant for deployment, rather than only on in-domain validation sets. Secondly, if LoRA or related mechanisms are used, it may be preferable to constrain adaptation to parts of the network that modulate global semantics while keeping early, texture-sensitive layers frozen, or to regularise the adapted parameters with explicit open-set constraints (e.g., penalising large shifts in separability on validation PAIs that are treated as proxies for unseen attacks). The present study does not prescribe a particular strategy, but it highlights that unconstrained adaptation can reduce, rather than increase, open-set robustness.

\subsection{Decision Layer and Calibration under Cross-Spectral Shift}

The cross-spectral analysis suggests that geometric separability in the embedding space does not always translate into good operating points. Several frozen models retain negative SRD on VSIA, indicating that bona fide and attacks occupy distinct regions of feature space even when D-EER(\%) and BPCER are poor. This disconnect points to limitations at the decision layer: a single global threshold, learned on NIR data, is misaligned with VIS score distributions.

In practical PAD systems, this argues for calibration mechanisms that are spectrum-aware, such as:
\begin{itemize}
    \item spectrum-conditional thresholds, where different operating points are learned for NIR and VIS (or, more generally, for different sensor–spectrum combinations);
    \item score normalisation schemes that map scores from different spectra into a common reference scale before thresholding;
    \item simple spectrum detectors that route samples to spectrum-specific PAD heads sharing a common backbone but separate calibration.
\end{itemize}
While the present work does not implement such strategies, the observed discrepancy between feature-space separability and score-space performance suggests that cross-spectral PAD is partly a calibration problem in addition to a representation-learning problem.

\subsection{Evaluation Protocols and Reporting Practices}

Finally, the results underscore the importance of evaluation protocols that explicitly reflect open-set conditions. Standard closed-set or semi-closed protocols can give a misleading impression of robustness, particularly when foundation models are involved. For PAD design, it is therefore essential that:
\begin{itemize}
    \item open-set protocols explicitly isolate unseen PAIs, unseen datasets and cross-spectral shifts, as done here;
    \item metrics are reported together with confidence intervals, so that both performance level and statistical stability are visible;
    \item conclusions about robustness are always qualified by the specific type of shift under which they were obtained.
\end{itemize}
In this sense, the present study is not only a negative result about current vision foundation models, but also a methodological recommendation: PAD algorithms should be designed and evaluated with explicit reference to the open-set axes along which they are expected to operate.

Overall, the findings suggest that future iris PAD designs based on foundation models should combine three elements: (i) PAD-aware representation learning that preserves artefact-sensitive cues, (ii) carefully constrained and evaluated adaptation mechanisms, and (iii) decision-layer calibration that takes spectrum and sensor variability into account. Without these elements, strong performance on closed or cross-dataset benchmarks cannot be taken as evidence of security in realistic open-set deployments.
\begin{table*}[t]
\centering
\caption{Extended validation using segmented iris inputs. One representative condition is evaluated for each protocol. Values are reported as mean $\pm$ 95\% confidence interval. Lower values indicate better performance. Best results for each metric and protocol are shown in bold.}
\label{tab:segmented_iris_validation}
\footnotesize
\setlength{\tabcolsep}{5.2pt}
\renewcommand{\arraystretch}{1.08}
\begin{tabular}{llccc}
\toprule
\multirow{2}{*}{\textbf{Model}} &
\multirow{2}{*}{\textbf{Metric}} &
\textbf{Protocol~1} &
\textbf{Protocol~2} &
\textbf{Protocol~3} \\
\cmidrule(lr){3-3}\cmidrule(lr){4-4}\cmidrule(lr){5-5}
& &
\textbf{Diseased} &
\textbf{LivDet17 Clarkson} &
\textbf{VSIA} \\
\midrule

\multirow{3}{*}{DINOv3 (frozen)}
& D-EER (\%) $\downarrow$        & 46.76$\pm$2.99 & 42.96$\pm$1.09 & 41.54$\pm$0.76 \\
& BPCER5 (\%) $\downarrow$  & 97.52$\pm$2.48 & 95.42$\pm$1.87 & \best{86.35\pm1.11} \\
& BPCER10 (\%) $\downarrow$ & 89.72$\pm$4.97 & 87.71$\pm$1.65 & 79.23$\pm$1.14 \\
\midrule

\multirow{3}{*}{DINOv3 (LoRA)}
& D-EER (\%) $\downarrow$        & 51.42$\pm$2.92 & 29.01$\pm$1.09 & 45.00$\pm$0.69 \\
& BPCER5 (\%) $\downarrow$  & 99.65$\pm$0.71 & 88.34$\pm$2.58 & 91.15$\pm$0.81 \\
& BPCER10 (\%) $\downarrow$ & 95.74$\pm$3.19 & 74.68$\pm$3.26 & 84.42$\pm$1.52 \\
\midrule

\multirow{3}{*}{OpenVision (frozen)}
& D-EER (\%) $\downarrow$        & 44.01$\pm$2.95 & 21.59$\pm$0.92 & 41.35$\pm$0.73 \\
& BPCER5 (\%) $\downarrow$  & 99.65$\pm$0.71 & 75.14$\pm$4.02 & 87.88$\pm$1.11 \\
& BPCER10 (\%) $\downarrow$ & 97.87$\pm$5.32 & 51.64$\pm$3.67 & \best{76.92\pm1.34} \\
\midrule

\multirow{3}{*}{OpenVision (LoRA)}
& D-EER (\%) $\downarrow$        & \best{41.84\pm3.30} & \best{19.75\pm0.89} & \best{39.85\pm0.80} \\
& BPCER5 (\%) $\downarrow$  & \best{91.49\pm3.90} & \best{58.22\pm3.74} & 90.19$\pm$1.25 \\
& BPCER10 (\%) $\downarrow$ & \best{84.40\pm4.26} & \best{40.06\pm3.16} & 82.31$\pm$1.60 \\
\bottomrule
\end{tabular}
\end{table*}
\section{Extended Experimental Validation}
\label{sec:extended_validation}

Sections~\ref{sec:results} and~\ref{sec:sep-analysis} analyse vision foundation models under three isolated open-set shifts: unseen presentation attack instruments, unseen datasets, and cross-spectral transfer. We now conduct additional targeted experiments to test whether the main conclusions persist under alternative input representations, stronger backbone adaptation, joint open-set shifts, and reverse cross-spectral transfer. These experiments are intended as focused validations rather than a second exhaustive benchmark over all models and protocols.

Unless otherwise stated, the same training, validation, preprocessing, and evaluation procedures described in Sections~\ref{sec:datasets} and~\ref{sec:results} are used. The extended experiments are therefore performed on representative models and operating conditions to keep the analysis concise while addressing the main remaining design questions. Performance is reported using D-EER(\%), BPCER at APCER~=~5\%, and BPCER at APCER~=~10\%, together with 95\% confidence intervals.

\subsection{Effect of Iris Localisation}
\label{sec:extended_segmentation}

The main experiments use periocular images rather than explicitly segmented iris regions. This choice ensures a consistent preprocessing pipeline across datasets and spectra, and avoids introducing segmentation errors as an additional source of variation. Since many conventional iris PAD pipelines operate on localised iris regions, we further verify whether the open-set failures observed in the main study are caused by the use of periocular inputs.

For this validation, iris segmentation is first performed using the open-source method of~\cite{Segmentation_Geethanjali}. Since the segmentation method is not error-free across the heterogeneous datasets used in this study, the resulting masks are inspected and clear segmentation failures are corrected manually. Approximately 13\% of the segmented samples required such correction. The final quality-checked segmented iris regions are then used as input to the same foundation-model evaluation pipeline. This experiment is not intended to benchmark iris segmentation algorithms; rather, it tests whether replacing periocular inputs with iris-localised inputs changes the conclusions drawn from the main experiments. The periocular results are already reported in Sections~\ref{sec:results} and~\ref{sec:sep-analysis}, while the segmented setting provides the direct comparison needed to assess the effect of iris localisation.

The experiment is performed using DINOv3 and OpenVision. DINOv3 is selected because it is central to the main analysis and feature-space discussion, while OpenVision is selected because it provides comparatively strong frozen performance in the original cross-spectral setting. These two models capture different behaviours observed in the main study while keeping the validation focused. One representative condition is selected from each main protocol: the diseased-eye category for Protocol~1, LivDet-Iris-2017 Clarkson for Protocol~2, and VSIA for Protocol~3. These settings respectively represent unseen PAI, cross-dataset, and NIR-to-VIS cross-spectral evaluation. For each backbone, both frozen and LoRA-adapted variants are evaluated using the same training and evaluation procedure as in the main study.

Table~\ref{tab:segmented_iris_validation} reports the segmented-input results. Iris localisation does not remove the main open-set failure modes. Under Protocol~1, both DINOv3 and OpenVision continue to produce high D-EER(\%) and BPCER values on the held-out diseased-eye category. DINOv3 obtains a D-EER(\%) of $46.76\%$ in the frozen setting and $51.42\%$ with LoRA, while OpenVision obtains $44.01\%$ and $41.84\%$, respectively. Thus, the difficulty of unseen PAI generalisation is not primarily due to the use of periocular images. Under Protocol~2, segmentation does not provide a consistent gain in cross-dataset transfer. OpenVision performs better than DINOv3 on LivDet-Iris-2017 Clarkson, particularly with LoRA, where it obtains a D-EER(\%) of $19.75\%$. However, the corresponding BPCER values remain substantial, especially at the stricter APCER operating point. Under Protocol~3, all segmented-input configurations still yield high D-EER(\%) values on VSIA, ranging from $39.85\%$ to $45.00\%$, with high BPCER at both operating points. Therefore, while iris localisation can change absolute error rates for specific model--protocol combinations, it does not alter the central conclusion that current foundation-model representations remain vulnerable to unseen PAIs and spectral shifts, even when evaluated on segmented iris inputs.
\begin{table*}[t]
\centering
\caption{Extended validation using full backbone fine-tuning. One representative condition is evaluated for each protocol. Values are reported as mean $\pm$ 95\% confidence interval. Lower values indicate better performance. Best results for each metric and protocol are shown in bold.}
\label{tab:full_finetuning_validation}
\footnotesize
\setlength{\tabcolsep}{5.2pt}
\renewcommand{\arraystretch}{1.08}
\begin{tabular}{llccc}
\toprule
\multirow{2}{*}{\textbf{Model}} &
\multirow{2}{*}{\textbf{Metric}} &
\textbf{Protocol~1} &
\textbf{Protocol~2} &
\textbf{Protocol~3} \\
\cmidrule(lr){3-3}\cmidrule(lr){4-4}\cmidrule(lr){5-5}
& &
\textbf{Diseased} &
\textbf{LivDet17 Clarkson} &
\textbf{VSIA} \\
\midrule

\multirow{3}{*}{DINOv3 (Full FT)}
& D-EER (\%) $\downarrow$                 & \best{50.00\pm2.84}  & 50.00$\pm$1.14  & \best{20.98\pm0.50} \\
& BPCER5 (\%) $\downarrow$     & \best{100.00\pm0.00} & 99.49$\pm$0.27  & \best{41.35\pm1.29} \\
& BPCER10 (\%) $\downarrow$    & \best{98.58\pm1.42}  & 98.08$\pm$0.60  & \best{41.35\pm1.29} \\
\midrule

\multirow{3}{*}{DINOv2 (Full FT)}
& D-EER (\%) $\downarrow$                 & \best{50.00\pm0.00}  & \best{49.82\pm1.16}  & 50.00$\pm$0.00 \\
& BPCER5 (\%) $\downarrow$     & \best{100.00\pm0.00} & \best{93.60\pm1.11}  & 100.00$\pm$0.00 \\
& BPCER10 (\%) $\downarrow$    & 100.00$\pm$0.00 & \best{87.71\pm1.92}  & 100.00$\pm$0.00 \\
\bottomrule
\end{tabular}
\end{table*}
\subsection{Effect of Full Backbone Fine-Tuning}
\label{sec:extended_full_finetuning}

The main experiments compare frozen foundation-model features with parameter-efficient adaptation using LoRA. While LoRA is useful for analysing the effect of a small number of task-adapted parameters, it does not determine whether stronger adaptation of the full backbone can overcome the observed failures. We therefore conduct an additional full fine-tuning experiment in which all backbone parameters are updated jointly with the PAD classification head.

This experiment is a targeted validation rather than a second full benchmark. DINOv3 is selected because it is central to the main analysis and feature-space discussion. DINOv2 is included as a second self-supervised transformer that shows strong behaviour in parts of the original cross-dataset evaluation, particularly with LoRA. The same representative conditions used in Section~\ref{sec:extended_segmentation} are evaluated: diseased eyes for Protocol~1, LivDet-Iris-2017 Clarkson for Protocol~2, and VSIA for Protocol~3. The training and validation partitions, preprocessing, classifier head, metrics, and early stopping procedure are kept consistent with the main experiments. The only change is the adaptation regime: the full backbone is fine-tuned using a lower learning rate and weight decay.

Table~\ref{tab:full_finetuning_validation} reports the full fine-tuning results. Under Protocol~1, both DINOv3 and DINOv2 remain close to chance-level D-EER(\%) on the held-out diseased-eye category, with very high BPCER values. For DINOv3, full fine-tuning gives a D-EER(\%) of $50.00\%$, compared with $50.00\%$ for the frozen setting and $39.26\%$ for LoRA on the corresponding pathology condition in Table~\ref{tab:prot13}. Thus, stronger adaptation does not remove the unseen-PAI failure and may over-specialise to the observed training PAIs.

The same trend is observed under Protocol~2. On LivDet-Iris-2017 Clarkson, DINOv3 with full fine-tuning obtains a D-EER(\%) of $50.00\%$, whereas the corresponding frozen and LoRA results in Table~\ref{tab:prot2} are $14.34\%$ and $2.32\%$, respectively. DINOv2 also degrades substantially, reaching $49.82\%$ D-EER(\%) compared with $18.01\%$ in the frozen setting and $4.81\%$ with LoRA. This indicates that full fine-tuning can reduce cross-dataset robustness, even when parameter-efficient adaptation is beneficial.

The cross-spectral setting is more model-dependent. For DINOv3, full fine-tuning improves NIR-to-VIS D-EER(\%) to $20.98\%$, compared with $42.27\%$ for the frozen model and $44.04\%$ for LoRA in Table~\ref{tab:prot13}. However, BPCER remains high at $41.35\%$ at both operating points, which is still unsuitable for reliable PAD operation. In contrast, DINOv2 collapses under full fine-tuning in the same setting, reaching $50.00\%$ D-EER(\%) and $100.00\%$ BPCER at both operating points.

Overall, full backbone fine-tuning does not provide a general solution to open-set iris PAD. It can partially improve one cross-spectral case, but it fails under unseen PAI and cross-dataset shifts and is not consistent across models. These results show that the observed failures are not merely due to weak adaptation or frozen representations; updating the full backbone changes the adaptation bias, but does not by itself solve the open-set robustness problem.
\subsection{Joint Cross-Dataset and Cross-PAI Evaluation}
\label{sec:extended_protocol4}
\begin{table*}[t]
\centering
\caption{Extended validation under Protocol~4, which combines cross-dataset and cross-PAI shifts. Each column denotes a held-out dataset--PAI pair. Values are reported as mean $\pm$ 95\% confidence interval. Lower values indicate better performance. Best results for each metric and held-out dataset--PAI pair are shown in bold.}
\label{tab:protocol4_validation}
\footnotesize
\setlength{\tabcolsep}{4.2pt}
\renewcommand{\arraystretch}{1.08}
\begin{tabular}{llcccc}
\toprule
\multirow{2}{*}{\textbf{Model}} &
\multirow{2}{*}{\textbf{Metric}} &
\multicolumn{4}{c}{\textbf{Held-out dataset--PAI pair}} \\
\cmidrule(lr){3-6}
& &
\textbf{CASIA / Synthetic} &
\textbf{LivDet17 Clarkson / Print} &
\textbf{NDCLD15 / TCL} &
\textbf{Warsaw / Diseased} \\
\midrule

\multirow{3}{*}{\makecell{DINOv3\\(frozen)}}
& D-EER (\%) $\downarrow$                 & \best{50.00\pm0.64} & 33.27$\pm$1.21 & 50.00$\pm$1.37 & 49.66$\pm$3.19 \\
& BPCER5 (\%) $\downarrow$     & \best{71.08\pm0.88} & 74.94$\pm$3.08 & 95.56$\pm$1.22 & 95.74$\pm$2.83 \\
& BPCER10 (\%) $\downarrow$    & \best{62.55\pm1.21} & 61.00$\pm$3.14 & 89.98$\pm$1.58 & 93.97$\pm$3.90 \\
\midrule

\multirow{3}{*}{\makecell{DINOv3\\(LoRA)}}
& D-EER (\%) $\downarrow$                 & \best{50.00\pm0.60} & \best{13.89\pm0.93} & \best{49.73\pm1.17} & \best{40.00\pm2.64} \\
& BPCER5 (\%) $\downarrow$     & 98.91$\pm$0.21 & \best{21.12\pm1.90} & \best{85.74\pm9.69} & \best{91.84\pm4.25} \\
& BPCER10 (\%) $\downarrow$    & 97.30$\pm$0.29 & \best{15.60\pm1.52} & \best{85.74\pm1.46} & \best{84.40\pm5.32} \\
\midrule

\multirow{3}{*}{\makecell{DINOv3\\(Full FT)}}
& D-EER (\%) $\downarrow$                 & \best{50.00\pm0.52} & 32.60$\pm$1.24 & 49.82$\pm$1.22 & 46.44$\pm$3.84 \\
& BPCER5 (\%) $\downarrow$     & 99.31$\pm$0.15 & 84.42$\pm$3.19 & 94.67$\pm$1.13 & 92.91$\pm$2.84 \\
& BPCER10 (\%) $\downarrow$    & 98.52$\pm$0.25 & 69.60$\pm$3.87 & 89.29$\pm$1.50 & 86.52$\pm$4.62 \\
\bottomrule
\end{tabular}
\end{table*}

The three main protocols isolate individual sources of open-set shift: unseen PAIs in Protocol~1, unseen datasets in Protocol~2, and cross-spectral transfer in Protocol~3. This isolation is useful for diagnosis, but practical deployment may involve multiple shifts simultaneously. A PAD system may, for example, encounter a new acquisition source together with a previously unseen attack type. To evaluate this more demanding setting, we introduce an additional joint open-set validation protocol, referred to as Protocol~4.

Protocol~4 combines cross-dataset and cross-PAI evaluation. For each run, one dataset and one PAI category are held out simultaneously. Training uses the remaining NIR data, excluding both the held-out dataset and all samples belonging to the held-out PAI category. Validation is drawn from the remaining training pool using the same split strategy as in the main experiments. Testing is performed only on bona fide samples from the held-out dataset and attack samples from the same held-out dataset belonging to the held-out PAI category. The test set therefore differs from the training set along two axes: the acquisition dataset is unseen, and the attack mechanism is also unseen.

We use four dataset-PAI combinations that are supported by the available data and cover the major PAI groups considered in Protocol~1: CASIA-IrisV4 with synthetic iris attacks, LivDet-Iris-2017 Clarkson with paper prints, NDCLD15 with textured contact lenses, and Warsaw-BioBase-Disease with diseased-eye samples. These combinations are not intended to enumerate all possible cases, but to provide representative joint-shift settings in which the held-out dataset contains the corresponding held-out attack category. Since Protocol~4 is more restrictive and computationally demanding than the isolated protocols, we evaluate DINOv3 only. DINOv3 is selected because it is central to the main analysis and is used consistently in the extended experiments. Three adaptation regimes are compared: frozen features with an MLP head, LoRA adaptation, and full backbone fine-tuning. This allows us to examine whether increased adaptation improves robustness under joint shift, or instead increases sensitivity to the training distribution.

Table~\ref{tab:protocol4_validation} reports the results. The joint-shift setting is difficult for most dataset-PAI pairs. For CASIA-IrisV4 with synthetic attacks, all three adaptation regimes remain at chance-level D-EER, with particularly high BPCER for LoRA and full fine-tuning. A similar pattern is observed for NDCLD15 with textured contact lenses, where frozen, LoRA, and full fine-tuning all remain close to $50\%$ D-EER(\%). The Warsaw-BioBase-Disease case also remains difficult, although LoRA reduces D-EER(\%) to $40.00\%$ compared with $49.66\%$ for the frozen model and $46.44\%$ for full fine-tuning. The only setting with a substantial improvement is LivDet-Iris-2017 Clarkson with paper prints. In this case, LoRA reduces D-EER(\%) from $33.27\%$ to $13.89\%$, whereas full fine-tuning gives $32.60\%$ and does not preserve the LoRA gain. This behaviour is consistent with the main Protocol~2 results, where LoRA can improve transfer when the held-out condition remains sufficiently related to the training distribution. However, this benefit is not consistent across joint-shift settings.

Overall, Protocol~4 shows that the failure modes identified under isolated shifts persist when dataset and PAI shifts occur jointly. LoRA can help in a specific dataset--PAI combination, but it does not provide reliable protection under joint open-set conditions. Full fine-tuning also does not solve the problem and often behaves similarly to, or worse than, the frozen baseline. These results support the need to evaluate iris PAD under simultaneous acquisition and attack-mechanism shifts, rather than relying only on isolated or semi-closed protocols.

\begin{table}[t]
\centering
\caption{Extended validation for reverse cross-spectral transfer. Models are trained on VIS data from VSIA and tested on the NIR corpus. Values are reported as mean $\pm$ 95\% confidence interval. Lower values indicate better performance. Best results for each metric are shown in bold.}
\label{tab:vis_to_nir_validation}
\footnotesize
\setlength{\tabcolsep}{4.2pt}
\renewcommand{\arraystretch}{1.08}
\begin{tabular}{llc}
\toprule
\textbf{Model} & \textbf{Metric} & \textbf{VIS$\rightarrow$NIR} \\
\midrule

\multirow{3}{*}{CLIP (frozen)}
& D-EER (\%) $\downarrow$                 & 50.00$\pm$0.38 \\
& BPCER5 (\%) $\downarrow$     & 96.98$\pm$0.27 \\
& BPCER10 (\%) $\downarrow$    & 92.56$\pm$0.39 \\
\midrule

\multirow{3}{*}{CLIP (LoRA)}
& D-EER (\%) $\downarrow$                 & 50.00$\pm$0.34 \\
& BPCER5 (\%) $\downarrow$     & 98.01$\pm$0.18 \\
& BPCER10 (\%) $\downarrow$    & 95.56$\pm$0.29 \\
\midrule

\multirow{3}{*}{EVA02 (frozen)}
& D-EER (\%) $\downarrow$                 & 50.00$\pm$0.40 \\
& BPCER5 (\%) $\downarrow$     & 81.73$\pm$0.39 \\
& BPCER10 (\%) $\downarrow$    & 81.73$\pm$0.39 \\
\midrule

\multirow{3}{*}{EVA02 (LoRA)}
& D-EER (\%) $\downarrow$                 & 50.00$\pm$0.44 \\
& BPCER5 (\%) $\downarrow$     & \best{75.23\pm0.47} \\
& BPCER10 (\%) $\downarrow$    & 75.23$\pm$0.47 \\
\midrule

\multirow{3}{*}{DINOv2 (frozen)}
& D-EER (\%) $\downarrow$                 & \best{32.95\pm0.31} \\
& BPCER5 (\%) $\downarrow$     & 83.01$\pm$0.99 \\
& BPCER10 (\%) $\downarrow$    & \best{67.25\pm0.78} \\
\midrule

\multirow{3}{*}{DINOv2 (LoRA)}
& D-EER (\%) $\downarrow$                 & 47.09$\pm$0.34 \\
& BPCER5 (\%) $\downarrow$     & 81.84$\pm$0.38 \\
& BPCER10 (\%) $\downarrow$    & 81.84$\pm$0.38 \\
\midrule

\multirow{3}{*}{DINOv3 (frozen)}
& D-EER (\%) $\downarrow$                 & 48.61$\pm$0.30 \\
& BPCER5 (\%) $\downarrow$     & 96.58$\pm$0.27 \\
& BPCER10 (\%) $\downarrow$    & 92.88$\pm$0.33 \\
\midrule

\multirow{3}{*}{DINOv3 (LoRA)}
& D-EER (\%) $\downarrow$                 & 50.00$\pm$0.35 \\
& BPCER5 (\%) $\downarrow$     & 96.54$\pm$0.21 \\
& BPCER10 (\%) $\downarrow$    & 93.10$\pm$0.33 \\
\midrule

\multirow{3}{*}{OpenVision (frozen)}
& D-EER (\%) $\downarrow$                 & 50.00$\pm$0.37 \\
& BPCER5 (\%) $\downarrow$     & 93.36$\pm$0.26 \\
& BPCER10 (\%) $\downarrow$    & 91.02$\pm$0.30 \\
\midrule

\multirow{3}{*}{OpenVision (LoRA)}
& D-EER (\%) $\downarrow$                 & 50.00$\pm$0.35 \\
& BPCER5 (\%) $\downarrow$     & 81.93$\pm$0.46 \\
& BPCER10 (\%) $\downarrow$    & 81.93$\pm$0.46 \\
\bottomrule
\end{tabular}
\end{table}
\subsection{Reverse Cross-Spectral Evaluation}
\label{sec:extended_reverse_spectral}

Protocol~3 evaluates cross-spectral transfer from NIR to VIS. This direction is operationally motivated because most deployed iris recognition systems acquire NIR images, while VIS imagery represents a distinct acquisition condition. To provide a more complete view of spectral transfer, we also evaluate the reverse direction by training on VIS data and testing on NIR data.

This experiment should be interpreted as a reverse cross-spectral validation under limited VIS-source support. Unlike the NIR training pool used in Protocol~3, which combines multiple datasets, sensors, and PAI categories, the VIS training data come only from VSIA. The objective is therefore not to claim symmetry between the two directions, but to examine whether the failure observed under NIR$\rightarrow$VIS transfer also appears when the source and target spectra are reversed. The models are trained using VSIA, including VIS bona fide samples and the available VIS attack classes, and are tested on the combined NIR corpus. Since the VIS and NIR corpora differ in spectrum, dataset diversity, and PAI composition, this setting includes both spectral and attack-distribution mismatch. In contrast to the preceding targeted experiments, all five foundation models are included because the experiment is compact and directly complements the original cross-spectral evaluation. For each model, both frozen and LoRA-adapted variants are evaluated using the same metric definitions as in the main study. Table~\ref{tab:vis_to_nir_validation} reports the VIS$\rightarrow$NIR results. The reverse direction also produces severe degradation for most models. CLIP, EVA02, DINOv3, and OpenVision operate close to chance-level D-EER(\%) in both frozen and LoRA settings. DINOv2 shows partial transfer in the frozen setting, with a D-EER(\%) of $32.95\%$, but its BPCER remains high, reaching $83.01\%$ at APCER~=~5\%. LoRA further degrades DINOv2, increasing the D-EER(\%) to $47.09\%$.

These results show that cross-spectral failure is not limited to the original NIR$\rightarrow$VIS direction. Even when the models are trained on VIS samples and evaluated on NIR imagery, spectral transfer remains unreliable, and high BPCER values indicate that strict PAD operating points are not achieved. This supports the conclusion that the main difficulty is not merely the choice of NIR as the source spectrum, but the lack of representations and calibration mechanisms that remain stable across spectral domains.

\section{Conclusion}

This work provides a systematic and protocol-controlled failure analysis of general-purpose vision foundation models for open-set iris PAD. Unlike prior transfer learning studies that primarily report improvements under closed-set, controlled, or limited cross-dataset settings, we examine three orthogonal open-set shifts: unseen presentation attack instruments, cross-dataset transfer, and cross-spectral NIR--VIS generalisation. Their impact is quantified using PAD operating-point metrics together with bootstrap confidence intervals.

The results reveal a critical discrepancy between current evaluation practice and deployment-relevant robustness. Foundation models can transfer across datasets under a fixed sensing modality, especially when limited parameter-efficient adaptation such as LoRA is used. However, this apparent robustness does not generalise to more operationally relevant cases. The evaluated models fail to generalise reliably to unseen presentation attack instruments and degrade sharply under spectral mismatch. In several cases, LoRA amplifies rather than mitigates these failures. The feature-space separability analysis further shows that the degradation is associated with changes in embedding geometry, including increased intra-class dispersion and reduced separability, rather than being only a consequence of operating-point misalignment.

These findings complement recent foundation-model-based iris PAD studies that report promising results under closed-set, controlled, or spectrally adapted evaluation settings~\cite{tapia2025towards,ramachandra2025spectrairispad}. In contrast to those works, the present study separates attack-level, dataset-level, and spectral shifts, and shows that favourable performance under one source of variation does not necessarily imply robustness under another. This distinction helps explain why general-purpose foundation-model representations may appear effective in controlled PAD settings while remaining unreliable under deployment-oriented open-set protocols.
The extended validation experiments further support this conclusion. Segmented iris inputs do not remove the observed open-set failures, indicating that the use of periocular imagery is not the primary cause of the degradation. Full backbone fine-tuning also fails to provide consistent robustness, particularly under unseen PAI and cross-dataset conditions. The joint cross-dataset and cross-PAI evaluation shows that isolated failure modes persist when shifts occur simultaneously, while the reverse VIS-to-NIR experiment confirms that cross-spectral transfer remains unreliable in both directions.

Overall, the study shows that large-scale pretraining does not inherently encode the artefact sensitivities required for robust iris PAD, despite providing domain-invariant cues that are useful for semantic and biometric recognition tasks. Success on cross-dataset benchmarks should therefore not be interpreted as evidence of open-set security. Future PAD representations must preserve sensitivity to artefact-level cues while remaining robust to nuisance variation from sensors, datasets, and spectra. This trade-off should be explicitly evaluated rather than treated as an emergent property of model scale or generic pretraining.


{
\small{
\bibliographystyle{IEEEtran}
\bibliography{Iris_PAD_Bib}

@article{morales2023introduction,
  title={Introduction to presentation attack detection in iris biometrics and recent advances},
  author={Morales, Aythami and Fierrez, Julian and Galbally, Javier and Gomez-Barrero, Marta},
  journal={Handbook of Biometric Anti-Spoofing: Presentation Attack Detection and Vulnerability Assessment},
  pages={103--121},
  year={2023},
  publisher={Springer}
}

@article{boyd2023comprehensive,
  title={Comprehensive study in open-set iris presentation attack detection},
  author={Boyd, Aidan and Speth, Jeremy and Parzianello, Lucas and Bowyer, Kevin W and Czajka, Adam},
  journal={IEEE Transactions on Information Forensics and Security},
  volume={18},
  pages={3238--3250},
  year={2023},
  publisher={IEEE}
}

@incollection{yambay2023review,
  title={Review of iris presentation attack detection competitions},
  author={Yambay, David and Das, Priyanka and Boyd, Aidan and McGrath, Joseph and Fang, Zhaoyuan and Czajka, Adam and Schuckers, Stephanie and Bowyer, Kevin and Vatsa, Mayank and Singh, Richa and others},
  booktitle={Handbook of Biometric Anti-Spoofing: Presentation Attack Detection and Vulnerability Assessment},
  pages={149--169},
  year={2023},
  publisher={Springer}
}

@article{nguyen2024deep,
  title={Deep learning for iris recognition: A survey},
  author={Nguyen, Kien and Proen{\c{c}}a, Hugo and Alonso-Fernandez, Fernando},
  journal={ACM Computing Surveys},
  volume={56},
  number={9},
  pages={1--35},
  year={2024},
  publisher={ACM New York, NY}
}

@article{ramachandra2025spectrairispad,
  title={SpectraIrisPAD: Leveraging Vision Foundation Models for Spectrally Conditioned Multispectral Iris Presentation Attack Detection},
  author={Ramachandra, Raghavendra and Venkatesh, Sushma},
  journal={IEEE Transactions on Biometrics, Behavior, and Identity Science (T-BIOM) },
  year={2025}
}

@article{choudhary2020iris,
  title={Iris anti-spoofing through score-level fusion of handcrafted and data-driven features},
  author={Choudhary, Meenakshi and Tiwari, Vivek and others},
  journal={Applied Soft Computing},
  volume={91},
  pages={106206},
  year={2020},
  publisher={Elsevier}
}

@article{raghavendra2015robust,
  title={Robust scheme for iris presentation attack detection using multiscale binarized statistical image features},
  author={Raghavendra, Ramachandra and Busch, Christoph},
  journal={IEEE Transactions on Information Forensics and Security},
  volume={10},
  number={4},
  pages={703--715},
  year={2015},
  publisher={IEEE}
}

@inproceedings{parzianello2022saliency,
  title={Saliency-guided textured contact lens-aware iris recognition},
  author={Parzianello, Lucas and Czajka, Adam},
  booktitle={Proceedings of the IEEE/CVF Winter Conference on Applications of Computer Vision},
  pages={330--337},
  year={2022}
}

@inproceedings{yadav2019detecting,
  title={Detecting textured contact lens in uncontrolled environment using DensePAD},
  author={Yadav, Daksha and Kohli, Naman and Vatsa, Mayank and Singh, Richa and Noore, Afzel},
  booktitle={Proceedings of the IEEE/CVF Conference on Computer Vision and Pattern Recognition Workshops},
  pages={0--0},
  year={2019}
}

@inproceedings{fang2020micro,
  title={Micro stripes analyses for iris presentation attack detection},
  author={Fang, Meiling and Damer, Naser and Kirchbuchner, Florian and Kuijper, Arjan},
  booktitle={2020 IEEE International Joint Conference on Biometrics (IJCB)},
  pages={1--10},
  year={2020},
  organization={IEEE}
}

@article{gautam2022deep,
  title={Deep supervised class encoding for iris presentation attack detection},
  author={Gautam, Gunjan and Raj, Aditya and Mukhopadhyay, Susanta},
  journal={Digital Signal Processing},
  volume={121},
  pages={103329},
  year={2022},
  publisher={Elsevier}
}

@article{balashanmugam2023effective,
  title={An effective model for the iris regional characteristics and classification using deep learning alex network},
  author={Balashanmugam, Thiyaneswaran and Sengottaiyan, Kumarganesh and Kulandairaj, Martin Sagayam and Dang, Hien},
  journal={IET Image Processing},
  volume={17},
  number={1},
  pages={227--238},
  year={2023},
  publisher={Wiley Online Library}
}

@article{sony2025benchmarking,
  title={Benchmarking Foundation Models for Zero-Shot Biometric Tasks},
  author={Sony, Redwan and Farmanifard, Parisa and Alzwairy, Hamzeh and Shukla, Nitish and Ross, Arun},
  journal={arXiv preprint arXiv:2505.24214},
  year={2025}
}

@article{shahreza2025foundation,
  title={Foundation models and biometrics: A survey and outlook},
  author={Shahreza, Hatef Otroshi and Marcel, S{\'e}bastien},
  journal={IEEE Transactions on Information Forensics and Security},
  year={2025},
  publisher={IEEE}
}

@article{tapia2025towards,
  title={Towards iris presentation attack detection with foundation models},
  author={Tapia, Juan E and Gonz{\'a}lez-Soler, L{\'a}zaro Janier and Busch, Christoph},
  journal={arXiv preprint arXiv:2501.06312},
  year={2025}
}

@article{radford2021learning,
  title={Learning transferable visual models from natural language supervision},
  author={Radford, Alec and Kim, Jong Wook and Hallacy, Chris and Ramesh, Aditya and Goh, Gabriel and Agarwal, Sandhini and Sastry, Girish and Askell, Amanda and Mishkin, Pamela and Clark, Jack and others},
  journal={Proceedings of the 38th International Conference on Machine Learning},
  year={2021}
}

@article{fang2024eva02,
  title={EVA-02: A visual representation for neon genesis},
  author={Fang, Yuxin and Sun, Qiang and Wang, Xiaofei and Huang, Tiejun and Wang, Xiaogang and Cao, Yue},
  journal={Image and Vision Computing},
  volume={149},
  pages={105171},
  year={2024}
}

@article{caron2024dinov3,
  title={DINOv3: Learning robust visual features without supervision},
  author={Caron, Mathilde and Touvron, Hugo and Misra, Ishan and others},
  journal={arXiv preprint arXiv:2404.07143},
  year={2024}
}

@article{oquab2023dinov2,
  title={DINOv2: Learning robust visual features without supervision},
  author={Oquab, Maxime and Darcet, Timoth{\'e}e and Moutakanni, Th{\'e}o and others},
  journal={Transactions on Machine Learning Research},
  year={2023}
}

@article{li2025openvision,
  title={OpenVision: A fully-open, cost-effective family of advanced vision encoders for multimodal learning},
  author={Li, Xianhang and others},
  journal={Proceedings of the IEEE/CVF International Conference on Computer Vision},
  year={2025}
}

@misc{casia_irisv4,
  title={CASIA Irisv4 Image Database},
  author={{Chinese Academy of Sciences' Institute of Automation (CASIA)}},
  howpublished={\url{http://www.cbsr.ia.ac.cn/china/Iris\%20Databases\%20CH.asp}}
}

@inproceedings{casia_iris_syn,
  title={Synthesis of large realistic iris databases using patch-based sampling},
  author={Wei, Zhuoshi and Tan, Tieniu and Sun, Zhenan},
  booktitle={2008 19th International Conference on Pattern Recognition},
  pages={1--4},
  year={2008},
  organization={IEEE}
}

@article{iiitd_cli,
  title={Unraveling the Effect of Textured Contact Lenses on Iris Recognition},
  author={Yadav, Daksha and Kohli, Naman and Doyle, James S. and Singh, Richa and Vatsa, Mayank and Bowyer, Kevin W.},
  journal={IEEE Transactions on Information Forensics and Security},
  volume={14},
  number={2},
  year={2019}
}

@inproceedings{livdet2013,
author = {Yambay, David and Doyle, James and Czajka, Adam and Bowyer, Kevin and Schuckers, Stephanie},
year = {2014},
month = {09},
pages = {},
title = {LivDet-Iris 2013 – Iris Liveness Detection Competition 2013},
journal = {IJCB 2014 - 2014 IEEE/IAPR International Joint Conference on Biometrics},
doi = {10.1109/BTAS.2014.6996283}
}

@inproceedings{livdet2015,
author = {Yambay, David and Walczak, Brian and Schuckers, Stephanie and Czajka, Adam},
year = {2017},
month = {02},
pages = {},
title = {LivDet-Iris 2015 – Iris Liveness Detection Competition 2015},
doi = {10.1109/ISBA.2017.7947701}
}

@inproceedings{livdet2017,
  title={LivDet-Iris 2017 -- Iris Liveness Detection Competition},
  author={Czajka, Adam and others},
  booktitle={IEEE International Joint Conference on Biometrics (IJCB)},
  year={2017}
}

@inproceedings{ndcld13,
  title={Variation in accuracy of textured contact lens detection based on sensor and lens pattern},
  author={Doyle, James S and Bowyer, Kevin W and Flynn, Patrick J},
  booktitle={2013 IEEE sixth international conference on biometrics: theory, applications and systems (BTAS)},
  pages={1--7},
  year={2013},
  organization={IEEE}
}

@article{ndcld15,
  title={Robust detection of textured contact lenses in iris recognition using BSIF},
  author={Doyle, James S and Bowyer, Kevin W},
  journal={IEEE Access},
  volume={3},
  pages={1672--1683},
  year={2015},
  publisher={IEEE}
}

@article{disease_pre_post,
  title={Eye Diseases and Their Impact on Iris Recognition Reliability},
  author={Trokielewicz, Mateusz and others},
  journal={IEEE Transactions on Information Forensics and Security},
  year={2016}
}

@inproceedings{warsaw-biobase,
  title={Assessment of iris recognition reliability for eyes affected by ocular pathologies},
  author={Trokielewicz, Mateusz and Czajka, Adam and Maciejewicz, Piotr},
  booktitle={2015 IEEE 7th International Conference on Biometrics Theory, Applications and Systems (BTAS)},
  pages={1--6},
  year={2015},
  organization={IEEE}
}

@inproceedings{synthetic_iris,
  title={Privacy-safe iris presentation attack detection},
  author={Mitcheff, Mahsa and Tinsley, Patrick and Czajka, Adam},
  booktitle={2024 IEEE International Joint Conference on Biometrics (IJCB)},
  pages={1--10},
  year={2024},
  organization={IEEE}
}

@inproceedings{hoffman2018convolutional,
  title={Convolutional neural networks for iris presentation attack detection: Toward cross-dataset and cross-sensor generalization},
  author={Hoffman, Steven and Sharma, Renu and Ross, Arun},
  booktitle={Proceedings of the IEEE conference on computer vision and pattern recognition workshops},
  pages={1620--1628},
  year={2018}
}

@article{bauer1998linear,
  title={The linear separability effect in color visual search: Ruling out the additive color hypothesis},
  author={Bauer, Ben and Jolicoeur, Pierre and Cowan, William B},
  journal={Perception \& Psychophysics},
  volume={60},
  number={6},
  pages={1083--1093},
  year={1998},
  publisher={Springer}
}

@INPROCEEDINGS{Segmentation_Geethanjali,
  author={Sharma, Geetanjali and Nagaich, Dev and Jaswal, Gaurav and Nigam, Aditya and Ramachandra, Raghavendra},
  booktitle={2025 IEEE International Joint Conference on Biometrics (IJCB)}, 
  title={VREyeSAM: Virtual Reality Non-Frontal Iris Segmentation using Foundational Model with uncertainty weighted loss}, 
  year={2025},
  volume={},
  number={},
  pages={1-9},
  doi={10.1109/IJCB65343.2025.11411123}}

@INPROCEEDINGS{Pal,
  author={Pal, Debasmita and Sony, Redwan and Ross, Arun},
  booktitle={2025 IEEE/CVF Winter Conference on Applications of Computer Vision (WACV)}, 
  title={A Parametric Approach to Adversarial Augmentation for Cross-Domain Iris Presentation Attack Detection}, 
  year={2025},
  volume={},
  number={},
  pages={5719-5729},
  doi={10.1109/WACV61041.2025.00558}}

@manual{ISO/IEC2015a,
	Author = {{ISO/IEC JTC1 SC37 Biometrics}},
	Organization = {International Organization for Standardization},
	Title = {{ISO/IEC} 30107-3. Information Technology - Biometric
	presentation attack detection - Part 3: Testing and Reporting},
	Year = {2017}
}
}
}
\end{document}